\documentclass[11pt]{article}
\usepackage[T1]{fontenc}
\usepackage[utf8]{inputenc}
\usepackage{amsmath,amsfonts,amsthm,amssymb}
\usepackage{setspace}
\usepackage{fullpage}
\usepackage{fancyhdr}
\usepackage{xspace}

\usepackage{booktabs}       %
\usepackage{nicefrac}       %
\usepackage{microtype}      %
\usepackage{xcolor}         %

\usepackage{verbatim}
\usepackage{lastpage}
\usepackage{extramarks}
\usepackage{natbib}
\usepackage{diagbox}

\usepackage{titletoc}

\usepackage{times}
\usepackage{graphicx,float,wrapfig}
\usepackage{algorithm}
\usepackage{enumitem}
\usepackage[noend]{algpseudocode}
\usepackage{subcaption}

\allowdisplaybreaks

\newtheorem{theorem}{Theorem}[section]
\newtheorem{lemma}[theorem]{Lemma}

\newtheorem{proposition}[theorem]{Proposition}

\theoremstyle{definition}

\date{}

\usepackage{amsmath,amsfonts,bm}

\newcommand*{\defeq}{\triangleq}

\def\1{\bm{1}}

\makeatletter
\newcommand{\ve}{\@ifnextchar\bgroup{\velong}{{\bm{e}}}}
\newcommand{\velong}[1]{{\bm{#1}}}
\makeatother

\DeclareMathAlphabet{\mathsfit}{\encodingdefault}{\sfdefault}{m}{sl}
\SetMathAlphabet{\mathsfit}{bold}{\encodingdefault}{\sfdefault}{bx}{n}

\def\calC{{\mathcal{C}}}

\def\calF{{\mathcal{F}}}

\def\calH{{\mathcal{H}}}

\def\calN{{\mathcal{N}}}

\def\calX{{\mathcal{X}}}

\newcommand{\E}{\mathbb{E}}
\newcommand{\R}{\mathbb{R}}

\newcommand{\Var}{\mathrm{Var}}

\DeclareMathOperator{\Tr}{Tr}

\def\({\left(}
\def\){\right)}
\def\[{\left[}
\def\]{\right]}

\newcommand{\diag}{\mathrm{diag}}

\newcommand{\Z}{\mathbb{Z}}

\newcommand{\relu}{\mathrm{ReLU}}

\newcommand{\dist}{\mathrm{dist}}

\newcommand{\abs}[1]{{\left| {#1} \right|}}

\DeclareMathOperator{\supp}{supp}
\newcommand{\ind}[1]{\mathbb{I}\left[ #1 \right]}

\newcommand{\erlzb}{\epsilon_{\calF}}
\newcommand{\lmin}{\lambda_{\rm min}}
\newcommand{\lmax}{\lambda_{\rm max}}

\newcommand{\dd}{{\rm d}}
\newcommand{\frer}{\calF\textup{-RER}}

\usepackage{hyperref}
\usepackage{url}

\title{First Steps Toward Understanding the Extrapolation of Nonlinear Models to Unseen Domains}

\author{Kefan Dong \\ 
	Stanford University \\
	\texttt{kefandong@stanford.edu}
	\and
	Tengyu Ma \\
	Stanford University \\
	\texttt{tengyuma@stanford.edu}
}

\begin{document}

\maketitle

\begin{abstract}
	Real-world machine learning applications often involve deploying neural networks to domains that are not seen in the training time. Hence, we need to understand the extrapolation of \textit{nonlinear} models---under what conditions on the distributions and function class, models can be guaranteed to extrapolate to new test distributions. The question is very challenging because even two-layer neural networks cannot be guaranteed to  extrapolate outside the support of the training distribution without further assumptions on the domain shift. This paper makes some initial steps toward analyzing the extrapolation of nonlinear models for structured domain shift. We primarily consider settings where the \textit{marginal} distribution of each coordinate of the data (or subset of coordinates) does not shift significantly across the training and test distributions, but the joint distribution may have a much bigger shift. We prove that the family of nonlinear models of the form $f(x)=\sum f_i(x_i)$, where $f_i$ is an \emph{arbitrary} function on the subset of features $x_i$, can extrapolate to unseen distributions, if the covariance of the features is well-conditioned. To the best of our knowledge, this is the first result that goes beyond linear models and the bounded density ratio assumption, even though the assumptions on the distribution shift and function class are stylized.
\end{abstract}

\section{Introduction}\label{sec:intro}

In real-world applications, machine learning models are often deployed on domains that are not seen in the training time \citep{koh2021wilds,beery2018recognition,zech2018variable}. 
For example, we may train machine learning models for medical diagnosis on data from hospitals in Europe and then deploy them to hospitals in Asia. 

Thus, we need to understand the extrapolation of models to new test distributions---how the model trained on one distribution behaves on another unseen distribution. 
This extrapolation of neural networks  is central to various robustness questions such as domain generalization~\citep{gulrajani2020search,ganin2016domain,peters2016causal} %
and adversarial robustness~\citep{goodfellow2014explaining,kurakin2018adversarial}, and 
also plays a critical role in nonlinear bandits and reinforcement learning where the distribution is constantly changing during training \citep{dong2021provable,agarwal2019reinforcement,lattimore2020bandit,sutton2018reinforcement}. 

This paper focuses on the following mathematical abstraction of this extrapolation question:
\begin{center}\textit{
	Under what conditions on the source distribution $P$, target distribution $Q$, and function class $\calF$ do we have that any functions $f, g \in \calF$ that  agree on $P$ are also guaranteed to agree on $Q$?}
\end{center}

Here we can measure the agreement of two functions on $P$ by the $\ell_2$ distance between $f$ and $g$ under distribution $P$, that is, $\|f-g\|_P \defeq \mathbb{E}_{x\sim P} [(f(x)-g(x))^2]^{1/2}$. 
The function $f$ can be thought of as the learned model,  $g$ as the ground-truth function, and thus $\|f-g\|_P$ as the error on the source distribution $P$. 

This question is well-understood for linear function class $\mathcal{F}$. Essentially,  if the covariance of $Q$ can be bounded from above by the covariance of $P$ (in any direction), then the error on $Q$ is guaranteed to be bounded by the error on $P$. We refer the reader to~\citet{lei2021near,mousavi2020minimax} and references therein for more recent advances along this line.

By contrast, theoretical results for extrapolation of \textit{nonlinear} models are rather limited. 
Classical results have long settled the case where $P$ and $Q$ have bounded density ratios~\citep{ben2014domain,sugiyama2007covariate}. Bounded density ratio implies that the support of $Q$ must be a subset of the support of $P$, and thus arguably these results do not capture the extrapolation behavior of models \textit{outside} the training domain. 

Without the bounded density ratio assumption,  there was limited prior \textit{positive} result for characterizing the extrapolation power of neural networks. \citet{ben2010theory} show that the model can extrapolate when the $\calH\Delta\calH$-distance between training and test distribution is small. However, it remains unclear for what distributions and function class, the $\calH\Delta\calH$-distance can be bounded.\footnote{In fact, the $\calH\Delta\calH$-distance likely cannot be bounded when the function class contains two-layer neural networks, and the supports of the training and test distributions do not overlap ---when there exists a function that can distinguish the source and target domain, the $\calH\Delta\calH$ divergence will be large.}
In general, the question is challenging partly because of the existence of the following strong impossibility result. 
As soon as the support of $Q$ is not contained in the support of $P$ (and they satisfy some non-degeneracy condition), it turns out that even two-layer neural networks cannot extrapolate---there are two-layer neural networks $f$ and $g$ 
that agree on $P$ perfectly but behave very differently on $Q$. (See Proposition~\ref{prop:lowerbound} for a formal statement.)

The impossibility result suggests that any positive results on the extrapolation of nonlinear models require more fine-grained structures on the relationship between $P$ and $Q$ (which are common in practice~\citep{koh2021wilds,sagawa2021wilds}) as well as the function class $\calF$. The structure in the domain shift between $P$ and $Q$ may also need to be compatible with the assumption on the function class $\calF$. This paper makes some first steps towards proving certain family of nonlinear models can extrapolate to a new test domain with structured shift.

We consider a setting where the joint distribution of the data does not have much overlap across $P$ and $Q$ (and thus bounded density ratio assumption does not hold), whereas the marginal distributions for each coordinate of the data  does overlap. Such a scenario may practically happen when the features (coordinates of the data) exhibit different correlations on the source and target distribution. 
For example, consider the task of predicting the probability of a lightning storm from basic meteorological information such as precipitation, temperature, etc. We learn models from some cities on the west coast of United States and deploy them to the east coast. 
In this case, the joint test distribution of the features may not necessarily have much overlap with the training distribution---correlation between precipitation and temperature could be vastly different across regions, e.g., the rainy season coincides with the winter's low temperature on the west coast, but not so much on the east coast. However, the individual feature's marginal distribution is much more likely to overlap between the source and target---the possible ranges of temperature on east and west coasts are similar.

Concretely, we assume that the features $x\in \R^{d_1+d_2}$ have Gaussian distributions and can be divided into two subsets $x_1\in \R^{d_1}$ and $x_2\in \R^{d_2}$ %
such that each set of feature $x_i$ ($i\in \{1,2\}$) has the same  marginal distributions on $P$ and $Q$. Moreover, we assume that $x_1$ and $x_2$ are not exactly correlated on $P$---the covariance of features $x$ on distribution $P$ has a strictly positive minimum eigenvalue.

As argued before, restricted assumptions on the function class $\mathcal{F}$ are still necessary (for almost any $P$ and $Q$ without the bounded density ratio property). 
Here, we assume that $\mathcal{F}$ consists of all functions of the form $f(x) = f_1(x_1) + f_2(x_2)$ for \textit{arbitrary} functions $f_1:\R^{d_1}\rightarrow \R$ and $f_2:\R^{d_2}\rightarrow \R$. The function class $\calF$ does not contain all two-layer neural networks (so that the impossibility result does not apply), but still consists of a rich set of functions where each subset of features independently contribute to the prediction with arbitrary nonlinear transformations. We show that under these assumptions, if any two models approximately agree on $P$, they must also approximately agree on $Q$---formally speaking, $\forall f,g\in\calF,\; \|f-g\|_Q\lesssim \|f-g\|_P$ (Theorem~\ref{thm:gaussian-multi}). On synthetic datasets, we empirically verify that this structured function class indeed achieves better extrapolation than unstructured function classes (Section~\ref{sec:experiments}).

We also prove a variant of the result above where we divide features vector $x\in\R^{d}$ into $d$ coordinate, denoted by $x=(x_1,\dots, x_d)$ where $x_i\in \R$. The function class consists of all combinations of \textit{nonlinear} transformations of $x_i$'s, that is, $\calF=\{\sum_{i=1}^{d} f_i(x_i)\}$. %
Assuming coordinates of $x$ are pairwise Gaussian and $x$ has a non-degenerate covariance matrix, the nonlinear model $f\in\calF$ can extrapolate to any distribution $Q$ that has the same marginals as $P$ (Theorem~\ref{thm:gaussian}).

\begin{figure}
	\centering
	\includegraphics[width=.8\linewidth]{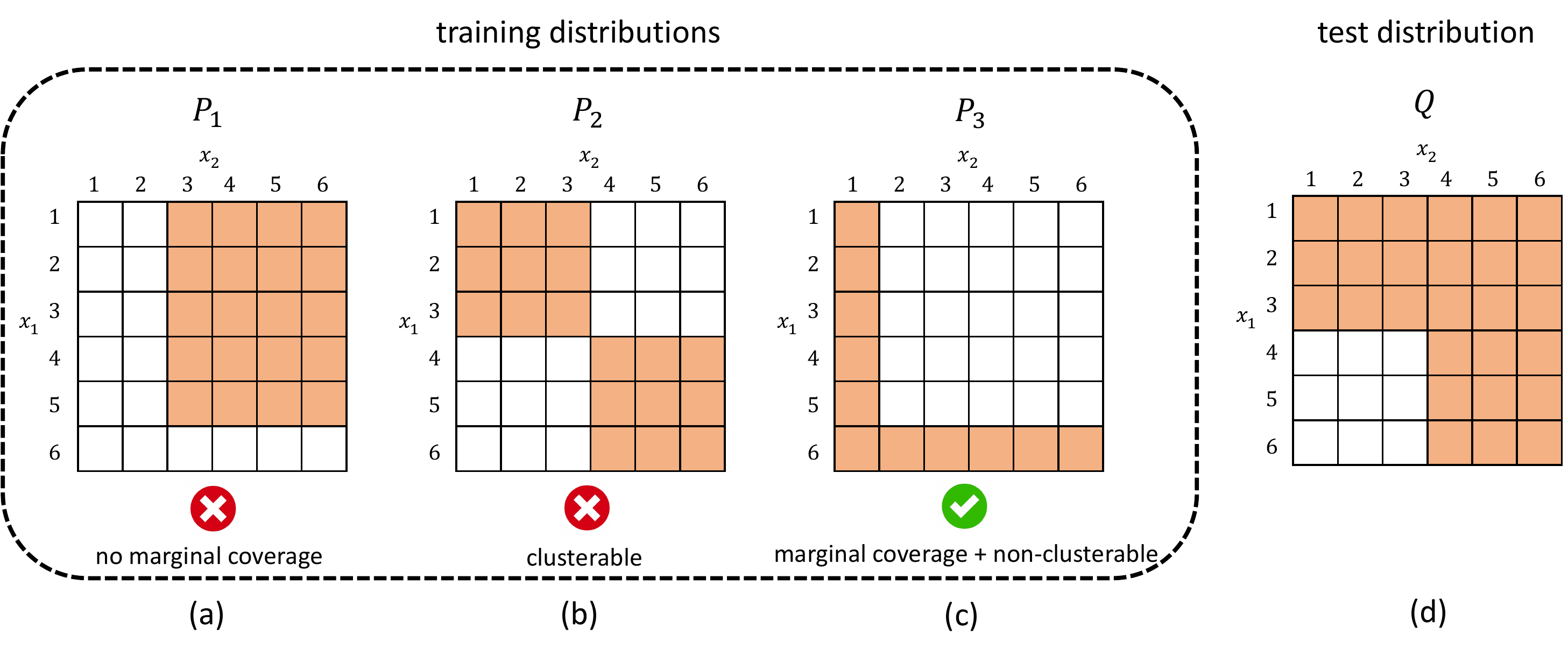}
	\caption{Visualization of three different training distributions $P_1,P_2,P_3$ and a test distribution $Q$, where the orange color blocks marks the support. 
		\textbf{(a)} and \textbf{(b):} distributions $P_1,P_2$ that do not satisfy our conditions. \textbf{(c):} a distribution $P_3$ that satisfies our conditions even though the support of $P_3$ is sparse.
	}
	\label{fig:1}
\end{figure}

These results can be viewed as first steps for analyzing extrapolation beyond linear models. Compared the works of~\cite{lei2021near} %
on linear models, our assumptions on the covariance of $P$ are qualitatively similar. We additionally require $P$ and $Q$ have overlapping marginal distributions because it is even necessary for the extrapolation of one-dimensional functions on a single coordinate. However, our results work for a more expressive family of nonlinear functions, that is, $\calF=\{\sum_{i=1}^{d} f_i(x_i)\}$, than the linear functions. 

We also present a result on the case where $x_i$'s are discrete variables, which demonstrates the key intuition and also may be of its own interest. 
Suppose we have two discrete random variables $x_1$ and $x_2$. In this case, the joint distribution of $P$ and $Q$ can be both presented by a matrix (as visualized in Figure~\ref{fig:1}), and the marginal distributions are the column and row sums of this joint probability matrix. 
We prove that extrapolation occurs when (1) the support of the \emph{marginals} of $Q$ is covered by $P$, and (2) the density matrix of $P$ is non-clusterable---we cannot shuffle the rows and columns of $P$ to form a block diagonal matrix where each block is a strict sub-matrix of $P$.

In Figure~\ref{fig:1}, we visualize a few interesting cases. First, distributions $P_1,P_2$ visualized in Figures 1a and 1b, respectively, do not satisfy our conditions. 
In contrast, our result predict that models trained on distribution $P_3$ in Figure 1c can extrapolate to the distribution $Q$ in Figure 1d, despite that the support of $P_3$ is sparse and the supports of $P,Q$ have very little overlap.

We also note that the failure of $P_1$ and $P_2$ demonstrate the non-triviality of our results. The overlapping marginal assumption by itself does not guarantees extrapolation, and condition (2), or analogously the minimal eigenvalue condition for the Gaussian cases,  is critical for extrapolation.

Our proof techniques start with viewing $\|f-g\|_P$ as $K_P(f-g,f-g)$ for some kernel $K_P:\calF\times\calF\to \R.$ 
Note that the kernel takes in two functions as inputs and captures relationship between the functions. 
Hence, the extrapolation of $\calF$ (i.e., proving $\|f-g\|_Q\lesssim \|f-g\|_P$ for all $f,g\in \calF$) reduces to the relationship of the kernels (i.e., whether $K_Q(f-g,f-g)\lesssim K_P(f-g,f-g)$ for all $f,g\in \calF$), which is then governed by properties of the eigenspaces of kernels $K_P,K_Q.$ Thanks to the special structure of our model class $\calF=\sum f_i(x_i)$, we can analytically relate the eigenspace of the kernels $K_P,K_Q$ to more explicitly and interpretable property of the data distribution of $P$ and $Q$.

\section{Problem Setup}\label{sec:setup}
We use $P$ and $Q$ to denote the source and target distribution over the space of features $\calX=\calX_1\times\cdots\times \calX_k$, respectively.
We measure the extrapolation of a model class $\calF\subseteq \R^{\calX}$ from $P$ to $Q$ by the $\mathcal{F}$-restricted error ratio, or $\mathcal{F}$-RER as a shorthand:\footnote{For simplicity, we set $0/0=0$.}
\begin{align}\label{equ:tau}
	\tau(P,Q,\calF)\defeq \sup_{f,g\in\calF}\frac{\E_Q[(f(x)-g(x))^2]}{\E_P[(f(x)-g(x))^2]}.
\end{align}
When $\tau(P, Q, \calF)$ is small, if two models $f,g\in\calF$ approximately agree on $P$ (meaning that $\E_P[(f(x)-g(x))^2]$ is small), they must approximately agree on $Q$ because $\E_Q[(f(x)-g(x))^2]\le \tau(P,Q,\calF) \E_P[(f(x)-g(x))^2].$ 

The $\mathcal{F}$-restricted error ratio monotonically increases as we enlarge the model class $\calF$, and eventually, $\tau(P, Q, \calF)$ becomes the density ratio between $Q$ and $P$ if the model $\calF$ contains all functions with bounded output.
To go beyond the bounded density ratio assumption, in this paper we focus on the structured model class $\calF=\{\sum_{i=1}^{k}f_i(x_i):\E_P[f_i(x_i)^2]<\infty,\forall i\in[k]\}$ where $f_i:\calX_i\to\R$ is an \emph{arbitrary} function.
Since $\calF$ is closed under addition, we can simplify Eq.~\eqref{equ:tau} to $\tau(P,Q,\calF)=\sup_{f\in\calF}\frac{\E_Q[f(x)^2]}{\E_P[f(x)^2]}.$ For simplicity, we omit the dependency on $P,Q,\calF$ when the context is clear. 

If the model class $\calF$ includes the ground-truth labeling function, $\tau(P,Q,\calF)$ upperbounds the ratio between the error on distribution $Q$ and the error on distribution $P$ (formally stated in Proposition~\ref{prop:losses}), which provides the robustness guarantee of the trained model. This is because when $g$ corresponds to the ground-truth label, $\E_P[(f(x)-g(x))^2]$ becomes the $\ell_2$ error of model $f$.
\begin{proposition}\label{prop:losses}%
	Let $\tau$ be the $\frer$ defined in Eq.~\eqref{equ:tau}. For any distribution $P,Q$ and model class $\calF$, if %
	there exists a model $f^\star$ in $\calF$ that can represent the true labeling $y:\calX\to\R$ on both $P$ and $Q$: %
	\begin{align}
		\E_{\frac{1}{2}(P+Q)}[(y(x)-f^\star(x))^2]\le \erlzb,
	\end{align}
	then we have
	\begin{align}
		\forall f\in\calF,\quad \E_{Q}[(y(x)-f(x))^2]\le (8\tau+4)\erlzb+4\tau \E_{P}[(y(x)-f(x))^2].
	\end{align}
\end{proposition}
Proof of this proposition is deferred to Appendix~\ref{app:pf-prop-losses}

\paragraph{Relationship to the $\calH\Delta \calH$-distance.} The $\frer$ is related to the $\calH\Delta \calH$-distance \citep{ben2010theory}:
\begin{align}
	\textstyle{d_{\calH\Delta\calH}(P,Q)=2\sup_{f,g\in\calF}\abs{\Pr_{x\sim P}[f(x)\neq g(x)]-\Pr_{x\sim Q}[f(x)\neq g(x)]},}
\end{align} with the differences that (1) we consider $\ell_2$ loss instead of classification loss, and (2) $\tau$ focuses on the ratio of losses whereas $d_{\calH\Delta\calH}$ focuses on the absolute difference. As we will see later, these differences bring the mathematical simplicity to prove concrete conditions for model extrapolation.

\paragraph{Additional notations.} Let $\ind{E}$ be the indicator function that equals $1$ if the condition $E$ is true, and $0$ otherwise. For an integer $n$, let $[n]$ be the set $\{1,2,\cdots,n\}$. For a vector $x\in\R^{d}$, we use $[x]_i$ to denote its $i$-th coordinate. Similarly, $[M]_{i,j}$ denotes the $(i,j)$-th element of a matrix $M$. We use $M^{\odot n}$ to represent the element-wise $n$-th power of the matrix $M$ (i.e., $[M^{\odot n}]_{i,j}=([M]_{i,j})^n$). 
Let $I_d\in\R^{d\times d}$ be the identity matrix, $\mathbf{1}_d\in\R^{d}$ the all-1 vector and $e_{i,d}$ the $i$-th base vector. We omit the subscript $d$ when the context is clear. For a square matrix $P\in\R^{d\times d}$, we use $\diag(P)\in\R^{d\times d}$ to denote the matrix generated by masking out all non-diagonal terms of $P$. For list $\sigma_1,\cdots,\sigma_d$, let $\diag(\{\sigma_1,\cdots,\sigma_d\})\in \R^{d\times d}$ be the diagonal matrix whose diagonal terms are $\sigma_1,\cdots,\sigma_d$.

For a symmetric matrix $M\in\R^{d\times d}$,let $\lambda_1(M)\le \lambda_2(M)\le \cdots \le \lambda_d(M)$ be its eigenvalues in ascending order, and $\lmax(M),\lmin(M)$ the maximum and minimum eigenvalue, respectively. Similarly, we use $\sigma_1(M), \cdots, \sigma_{\min(d_1,d_2)}(M)$ to denote the singular values of $M\in\R^{d_1\times d_2}$.

\section{Main Results}\label{sec:results}
In this section, we present our main results. Section~\ref{sec:two-features-discrete} discusses the case where the features have discrete values. 
In Section~\ref{sec:multi-features-Gaussian} and ~\ref{sec:vect-features-Gaussian}, we extend our analysis to two other settings with real-valued features.

\subsection{Features with Discrete Values}\label{sec:two-features-discrete}

For better exposition, we discuss the case that $x=(x_1,x_2)$ here and defer the discussion of the general case to Appendix~\ref{app:thm-discrete-multi}. We assume that $x_i$ takes the value in $\{1,2, \cdots,r_i\}$ for $i\in[2]$. Hence, the density of distribution $P$ can be written in a matrix with dimension $r_1\times r_2$. 

We measure the (approximate) clusterablity by eigenvalues of the Laplacian matrix of a bipartite graph associated with the density matrix $P\in\R^{r_1\times r_2}$, which is known to capture the clustering structure of the graph \citep{chung1996laplacians,alon1986eigenvalues}.
Let $G_P$ be a weighted bipartite graph whose adjacency matrix equals the density matrix $P\in\R^{r_1\times r_2}$---concretely,  $U=\{u_1,\cdots,u_{r_1}\}$ and $V=\{v_1,\cdots,v_{r_2}\}$ are the sets of vertices, and the weight between $u_i,v_j$ is $P(x_1=i,x_2=j)$.
To define the signless Laplacian of $G_P$, let $d_1\in\R^{r_1}$ and $d_2\in\R^{r_2}$ be the row and column sums of the weight matrix $P$ (in other words, degree of the vertices), and $D_1=\diag(d_1)\in\R^{r_1\times r_1},D_2=\diag(d_2)\in\R^{r_2\times r_2}$ the diagonal matrices induced by $d_1,d_2$, respectively. The signless Laplacian $K_P$ and normalized signless Laplacian $\bar{K}_P$ are:
\begin{align}\label{equ:discrete-kernel}
	K_P=\begin{pmatrix}
		D_1 & P\\
		P^\top & D_2
	\end{pmatrix},\quad \bar{K}_P=\diag(K_P)^{-1/2}K_P\diag(K_P)^{-1/2}.
\end{align}
Compared with the standard Laplacian matrix, the non-diagonal terms in the signless Laplacian $K_P$ are positive and equal to the absolute value of corresponding terms in the standard Laplacian matrix. In the following theorem, we bound the $\mathcal{F}$-RER from above by the eigenvalues of $\bar{K}_P$ and the density ratio of the marginal distributions of $Q,P$.

\begin{theorem}\label{thm:discrete}
	For any distributions $P,Q$ over discrete random variables $x_1,x_2$, and the model class $\calF=\{f_1(x_1)+f_2(x_2)\}$ where $f_i:\calX_i\to \R$ is an arbitrary function, the $\frer$ can be bounded above:
	\begin{align}\label{equ:twovar-discrete-main}
		\tau(P,Q,\calF)\le 2\lambda_2(\bar{K}_P)^{-1}\max_{i\in[2],t\in[r_i]}\frac{Q(x_i=t)}{P(x_i=t)}.
	\end{align}
\end{theorem}
Compared with prior works that assumes a bounded density ratio on the entire distribution (e.g., \citet{ben2014domain,sugiyama2007covariate}), we only require a bounded density ratio of the marginal distributions. 
In other words, the model class $f(x)=f_1(x_1)+f_2(x_2)$ can extrapolate to distributions with a larger support (see Figure~\ref{fig:1}c). In contrast, for an unstructured model (i.e., $f(x)$ is an arbitrary function of the entire input $x$), the model can behave arbitrarily on data points outside the support of $P$.

Qualitatively, Theorem~\ref{thm:discrete} proves sufficient conditions under which the structured model class can extrapolate to unseen distributions (as visualized in Figure~\ref{fig:1})---In particulary, Theorem~\ref{thm:discrete} implies that for non-trivial extrapolation, that is, $\tau(P,Q,\calF)<\infty$, we need (a) $\max_{i\in[2],t\in[r_i]}\frac{Q(x_i=t)}{P(x_i=t)}<\infty$ (i.e., the support of the \emph{marginals} of $Q$ is covered by $P$), and (b) $\lambda_2(\bar{K}_P)>0$. To interpret condition (b), note that Cheeger's inequality implies that $\lambda_2(\bar{K}_P)>0$ if and only if the bipartite graph $G_P$ is connected \citep{chung1996laplacians,alon1986eigenvalues}\footnote{Cheeger's inequality measures the clustering structure of a graph by the eigenvalues of its standard Laplacian. However, the signless Laplacian and standard Laplacian have the same eigenvalues for bipartite graphs \citet{cvetkovic2007signless,grone1990laplacian}.}, that is, there does not exist non-empty strict subsets of vertices $U'\subset U,V'\subset V$, such that $P(x_1\in U', x_2\not\in V')=0$ and $P(x_1\not\in U', x_2\in V')=0$. Equivalent, we cannot shuffle the rows and columns of $P$ to form a block diagonal matrix where each block is a strict sub-matrix of $P$. In other words, the density matrix $P$ is non-clusterable as discussed in Section~\ref{sec:intro}.

\paragraph{Proof sketch of Theorem~\ref{thm:discrete}.} In the following we present a proof sketch of Theorem~\ref{thm:discrete}. We start with a high-level proof strategy and then instantiate the proof on the setting of Theorem~\ref{thm:discrete}.

Suppose we can find a set of (not necessarily orthogonal) basis $\{b_1,\cdots,b_r\}$ where $b_i:\calX\to \R$, such that any model $f\in\calF$ can be represented as a linear combination of basis, that is, $f=\sum_{i=1}^{r}v_ib_i.$ Since the model family $\calF$ is closed under subtraction, we have
\begin{align}
	\tau(P,Q,\calF)&=\sup_{f\in\calF}\frac{\|f\|_Q^2}{\|f\|_P^2}
	=\sup_{v\in\R^{r}}\frac{\E_Q\big[\big(\sum_{i=1}^{r}v_ib_i(x)\big)^2\big]}{\E_P\big[\big(\sum_{i=1}^{r}v_ib_i(x)\big)^2\big]}=\sup_{v\in\R^{r}}\frac{\sum_{i,j=1}^{r}[v]_i[v]_j\E_Q[b_i(x)b_j(x)]}{\sum_{i,j=1}^{r}[v]_i[v]_j\E_P[b_i(x)b_j(x)]}.\label{equ:pfs-thm-d-1}
\end{align}
If we define the kernel matrices $[K_P]_{i,j}=\E_P[b_i(x)b_j(x)]$ and $[K_Q]_{i,j}=\E_Q[b_i(x)b_j(x)]$ (we use the same notation for the kernel matrix and signless Laplacian because later we will show that the kernels $K_P,K_Q$ equal to the signless Laplacian of the bipartite graphs $G_P,G_Q$ with specific choice of the basis $b_i$)
, it follows that 
\begin{align}
	\sup_{v\in\R^{r}}\frac{\sum_{i,j=1}^{r}v_iv_j\E_Q[b_i(x)b_j(x)]}{\sum_{i,j=1}^{r}v_iv_j\E_P[b_i(x)b_j(x)]}=\sup_{v\in\R^{r}}\frac{v^\top K_Q v}{v^\top K_P v}.
\end{align}
Hence, upper bounding $\tau(P,Q,\calF)$ reduces to bounding the eigenvalues of kernel matrices $K_P,K_Q$.

Since the model has the structure $f(x)=f_1(x_1)+f_2(x_2)$, we can construct the basis $\{b_t\}_{t=1}^{r}$ explicitly. For any $i\in[2],t\in[r_i]$, with little abuse of notation, let
\begin{align}
	b_{i,t}(x)=\ind{x_i=t}.
\end{align}
We can verify that the set $\{b_{i,t}\}_{i\in[2],t\in[r_i]}$ is indeed a complete set of basis.
As a result, the kernel matrices $K_P$ can be computed directly using its definition:
\begin{align}
	\E_{P}[b_{i,t}(x)b_{j,s}(x)]=\begin{cases}
		P(x_i=t,x_j=s),&\text{ when }i\neq j,\\
		P(x_i=t)\ind{s=t},&\text{ when }i= j,
	\end{cases}
\end{align}
which is exactly the Laplacian matrix defined in Eq.~\eqref{equ:discrete-kernel}.

To prove Eq.~\eqref{equ:twovar-discrete-main}, we need to upperbound the eigenvalues of $K_Q$. Since the eigenvalues of the normalized signless Laplacian $\bar{K}_Q$ is universally upper bounded by $2$ for every distribution $Q$, we first write $K_P,K_Q$ in terms of $\bar{K}_P,\bar{K}_Q$.
Formally, let $D_P=\diag(K_P)$ and $D_Q=\diag(K_Q)$ and we have
\begin{align}
	&\sup_{v\in\R^{r}}\frac{v^\top K_Qv}{v^\top K_P v}=\sup_{v\in\R^{r}}\frac{v^\top D_Q^{1/2} \bar{K}_Q D_Q^{1/2}v}{v^\top D_P^{1/2} \bar{K}_P D_P^{1/2}v}
	\le\frac{\lmax(\bar{K}_Q)}{\lmin(\bar{K}_P)}\sup_{v\in\R^{r}}\frac{\|D_Q^{1/2}v\|_2^2}{\|D_P^{1/2}v\|_2^2}.\label{equ:pf-sketch-1}
\end{align}
However, this naive bound is vacuous because for any $P$ we have $\lmin(\bar{K}_P)=0$. In fact, $K_P$ and $K_Q$ share the eigenvalue $0$ and the corresponding eigenvector $u\in\R^{r_1+r_2}$ with $[u]_t=(-1)^{\ind{t> r_1}}$. Therefore we can restrict to the subspace orthogonal to the direction $u$, and then $\lambda_{\min}(\bar{K}_P)$ becomes $\lambda_2(\bar{K}_P)$ in Eq.~\eqref{equ:pf-sketch-1}. Finally, by basic algebra we also have $\lmax(\bar{K}_Q)\le 2$ and $\sup_{v\in\R^{r}}\frac{\|D_Q^{1/2}v\|_2^2}{\|D_P^{1/2}v\|_2^2}\le \max_{i\in[2],t\in[r_i]}\frac{Q(x_i=t)}{P(x_i=t)}$, which complete the proof sketch. The full proof of Theorem~\ref{thm:discrete} is deferred to Appendix~\ref{app:pf-thm-discrete}.

\subsection{Features with Real Values}\label{sec:multi-features-Gaussian}
In this section we extend our analysis to the case where $x_1,x_2,\cdots,x_d$ are real-valued random  variables. Recall that our model has the structure $f(x)=\sum_{i=1}^{d} f_i(x_i)$ where $f_i$ is an arbitrary one-dimensional function. 

When $d=2$, we can view this setting as a direct extension of the setting in Section~\ref{sec:two-features-discrete} where each $x_i$'s has infinite number of possible values (instead of finite number), and thus the Laplacian ``matrix'' becomes infinite-dimensional. 
Nonetheless, we can still bound the $\mathcal{F}$-RER from above, as stated in the following theorem. 
\begin{theorem}\label{thm:gaussian}
	For any distributions $P,Q$ over variables $x=(x_1,\cdots,x_d)$ with matching marginals,
	assume that $(x_i,x_j)$ has the distribution of a two-dimensional Gaussian random variable for every $i,j\in[d]$. Let $\tilde{x}=(\tilde{x}_1,\cdots,\tilde{x}_d)$ be the normalized input where $\tilde{x}_i\defeq (x_i-\E_P[x_i])\Var(x_i)^{-1/2}$ has zero mean and unit variance for every $i\in[d]$, and $\tilde{\Sigma}_P\defeq \E_P[\tilde{x}\tilde{x}^\top]$ the covariance matrix of $\tilde{x}$. Then we have
	\begin{align}\label{equ:thm-gaussian-1}
		\tau(P,Q,\calF)\le \frac{d}{\lmin(\tilde{\Sigma}_P)}.
	\end{align}
\end{theorem}

For better exposition, we first focus on the case where every $x_i$ has zero mean and unit variance, hence $\tilde{x}=x$ and $\tilde{\Sigma}_P=\Sigma_P\defeq \E_P[xx^\top].$ Compared with linear models, Theorem~\ref{thm:gaussian} proves that the structured nonlinear model class $f(x)=\sum_{i=1}^{d}f_i(x_i)$ can extrapolate with similar conditions---for linear models $\calF_{\rm linear}\defeq \{v^\top x: v\in\R^{d}\}$ we have 
$$\tau(P,Q,\calF_{\rm linear})=\sup_{v\in\R^{d}}\frac{\|v^\top x\|_Q^2}{\|v^\top x\|_P^2}=\sup_{v\in\R^{d}}\frac{v^\top \E_Q[xx^\top] v}{v^\top \E_P[xx^\top] v}\lesssim \lmin(\Sigma_P)^{-1}=\lmin(\tilde{\Sigma}_P)^{-1},$$
which equals to the RHS of Eq.~\eqref{equ:thm-gaussian-1} up to factors that does not depend on the covariance $\Sigma_P,\Sigma_Q$.

We emphasize that we only assume the marginals on every \emph{pair} of features $x_i,x_j$ is Gaussian, which does not imply the Gaussianity of the joint distribution of $x$. In fact, there exists a non-Gaussian distribution that satisfies our assumption.

\paragraph{Proof sketch of Theorem~\ref{thm:gaussian}.} 
On a high level, we treat the features $x_i$'s as discrete random variables, and follow the same proof strategy as in Theorem~\ref{thm:discrete}. For better exposition, we first assume that $x_i$ has zero mean and unit variance for every $i\in[d],$ hence $\tilde{\Sigma}_P=\Sigma_P\defeq \E_P[xx\top].$

First we consider a simplified case when $d=2$.
Because $x_1, x_2$ are continuous, the normalized signless Laplacian $\bar{K}_P$ is infinite dimensional, and has the form $
	\bar{K}_P=\begin{pmatrix}
		I & A\\
		A^\top & I
	\end{pmatrix},
$ where $A$ is an infinite dimensional ``matrix'' indexed by real numbers $x_1,x_2\in\R$ with values $[A]_{x_1,x_2}=P(x_1,x_2)/\sqrt{P(x_1)P(x_2)}$, and $I$ is the identity ``matrix''.
Recall that in the proof of Theorem~\ref{thm:discrete} we get 
\begin{align}
	\tau(P,Q,\calF)\le  2\lambda_2(\bar{K}_P)^{-1}\max_{i\in[2],t}\frac{Q(x_i=t)}{P(x_i=t)}.
\end{align}
By the assumption that $P,Q$ have matching marginals, we get $\max_{i\in[2],t}\frac{Q(x_i=t)}{P(x_i=t)}=1$. As result, we only need to lowerbound the second smallest eigenvalue of $\bar{K}_P.$
To this end, we first write $A$ in its singular value decomposition form $A=U\Lambda V^\top$, where $UU^\top=I,VV^\top=I$ and $\Lambda=\diag(\{\sigma_n\}_{n\ge 0})$ with $\sigma_0\ge \sigma_1\ge \cdots$. Then we get
\begin{align}
	\bar{K}_P=\begin{pmatrix}
		I & A\\
		A^\top & I
	\end{pmatrix}=
	\begin{pmatrix}
		U & 0\\
		0 & V
	\end{pmatrix}
	\begin{pmatrix}
		I & \Lambda\\
		\Lambda^\top & I
	\end{pmatrix}
	\begin{pmatrix}
	U^\top & 0\\
	0 & V^\top
	\end{pmatrix}.
\end{align}
Since the matrix $\hat{K}_P\defeq \begin{pmatrix}
	I & \Lambda\\
	\Lambda^\top & I
\end{pmatrix}$ consists of four diagonal sub-matrices, we can shuffle the rows and columns of $\hat{K}_P$ to form a block-diagonal matrix with blocks $\left\{\begin{pmatrix}
1 & \sigma_n\\
\sigma_n & 1
\end{pmatrix}\right\}_{n=0,1,2,\cdots}.$
As a result, the eigenvalues of $\hat{K}_P$ are $1\pm \sigma_0,1\pm \sigma_1,\cdots$. Because $1=\sigma_0\ge \sigma_1\ge \cdots\ge 0$, the smallest and second smallest eigenvalues of $\hat{K}_P$ are $1-\sigma_0$ and $1-\sigma_1$, respectively, meaning that
$\lambda_2(\bar{K}_P)=\lambda_2(\hat{K}_P)=1-\sigma_1.$ By the assumption that $(x_1,x_2)$ follows from Gaussian distribution, the ``matrix'' $A$ is a Gaussian kernel, whose eigenvalues and eigenfunctions can be computed analytically---Theorem~\ref{thm:gaussian-kernel} proves that $\sigma_1=|\E_{P}[x_1x_2]|$ if $x_1,x_2$ have zero mean and unit variance. Consequently, $\lambda_2(\bar{K}_P)=1-\sigma_1=1-|\E_{P}[x_1x_2]|=\lmin(\Sigma_P).$

Now we briefly discuss the case when $d=3$, and the most general cases (i.e., $d>3$) are proved similarly. When $d=3$, the normalized kernel will have the following form
\begin{align}
	\bar{K}_P=\begin{pmatrix}
		I & A & B\\
		A^\top & I & C \\
		B^\top &C^\top & I
	\end{pmatrix}.
\end{align}
By the assumption that $x_1,x_2$ are zero mean and unit variance with joint Gaussian distribution, matrices $A$ is symmetric because $[A]_{x_1,x_2}=P(x_1,x_2)/\sqrt{P(x_1)P(x_2)}=P(x_2,x_1)/\sqrt{P(x_1)P(x_2)}=[A]_{x_2,x_1}$ Similarly, matrices $B,C$ are symmetric. In addition, Theorem~\ref{thm:gaussian-kernel} shows that the eigenfunctions of the Gaussian kernel is \emph{independent} of the value $\E_P[x_ix_j]$. Hence, the matrices $A,B,C$ shares the same eigenspace and can be diagonalized simultaneously:
\begin{align}
	\bar{K}_P=\begin{pmatrix}
		I & A & B\\
		A^\top & I & C \\
		B^\top &C^\top & I
	\end{pmatrix}=
	\begin{pmatrix}
		U & 0 & 0\\
		0 & U & 0\\
		0 & 0 & U
	\end{pmatrix}
	\begin{pmatrix}
		I & \Lambda_A & \Lambda_B\\
		\Lambda_A^\top & I & \Lambda_C\\
		\Lambda_B^\top & \Lambda_C^\top & I
	\end{pmatrix}
	\begin{pmatrix}
		U^\top & 0 & 0\\
		0 & U^\top & 0\\
		0 & 0 & U^\top
	\end{pmatrix}.
\end{align}
By reshuffling the columns and rows, the eigenvalues of $\bar{K}_P$ are the union of the eigenvalues of following matrices
\begin{align}
	\{\hat{K}_P^{(n)}\}_{n=0,1,2,\cdots}\defeq \left\{\begin{pmatrix}
		1 & \sigma_{n}(A) & \sigma_{n}(B)\\
		\sigma_{n}(A) & 1 & \sigma_{n}(C) \\
		\sigma_{n}(B) & \sigma_{n}(C) & 1
	\end{pmatrix}\right\}_{n=0,1,2,\cdots}.
\end{align}
Theorem~\ref{thm:gaussian-kernel} implies that $\sigma_{n}(A)=([\Sigma_P]_{1,2})^n,\sigma_{n}(B)=([\Sigma_P]_{1,3})^n$ and $\sigma_{n}(C)=([\Sigma_P]_{2,3})^n$. Consequently we get $\hat{K}_P^{(n)}=\Sigma_P^{\odot n}$. Then, this theorem follows directly by noticing $\lmin(\Sigma_P^{\odot n})\ge \lmin(\Sigma_P)$ for $n\ge 1$ (Lemma~\ref{lem:elementwise-product-lmin}).

Finally, the general case where $x_i$ has arbitrary mean and variance can be reduced to the case where $x_i$ has zero mean and unit variance (Lemma~\ref{lem:normalize}). The full proof of Theorem~\ref{thm:gaussian-kernel} is deferred to Appendix~\ref{app:pf-thm-gaussian}.

\subsection{Two Features with Multi-dimensional Gaussian Distribution}\label{sec:vect-features-Gaussian}
Now we extend Theorem~\ref{thm:gaussian} to the case where $x_1\in\R^{d_1},x_2\in\R^{d_2}$ are two subset of features with dimensions $d_1,d_2>1$, respectively, and the input $x=(x_1,x_2)$ has Gaussian distribution.
Recall that the model class is $\calF=\{f_1(x_1)+f_2(x_2):\E_P[f_i(x_i)^2]<\infty,\forall i\in[2]\}.$
In this case, we can still upper bound the $\frer$ by the eigenvalues of the covariance matrix $\Sigma_P$, which is stated in the following theorem.
\begin{theorem}\label{thm:gaussian-multi}
	For any distributions $P,Q$ over variables $x=(x_1,x_2)$ where $x_1\in\R^{d_1},x_2\in\R^{d_2}$, let $\Sigma_P=\E_P[xx^\top]$.
	If $x=(x_1,x_2)$ has Gaussian distribution on both $P$ and $Q$ with zero mean and matching marginals, and $\E_P[x_1x_1^\top]=I,\E_P[x_2x_2^\top]=I$, then
	\begin{align}\label{equ:thm-gm-1}
		\tau(P,Q,\calF)\le \frac{2}{\lmin(\Sigma_P)}.
	\end{align}
\end{theorem}
Compared with Theorem~\ref{thm:gaussian} where the features $x_1,\cdots,x_d$ are one-dimensional, our condition for the covariance is almost the same: $\lmin(\Sigma_P)>0$. However, the model class considered Theorem~\ref{thm:gaussian-multi} is more powerful because it captures nonlinear interactions between features within the same subset.
As a compromise, the assumption on the marginals of $P$ and $Q$ is stronger because Theorem~\ref{thm:gaussian-multi} requires matching marginals on each subset of the features, whereas Theorem~\ref{thm:gaussian} only requires matching marginals on each individual feature.

In the following, we show the proof sketch of Theorem~\ref{thm:gaussian-multi}, and the full proof is deferred to Appendix~\ref{app:pf-thm-gaussian-multi}.

\paragraph{Proof sketch of Theorem~\ref{thm:gaussian-multi}.} We start by considering a simpler case when $d_1=d_2$ and $\Sigma_{12}\defeq \E_P[x_1x_2^\top]\in\R^{d_1\times d_2}$ has only diagonal terms. In other words, the two dimensional random variable $([x_1]_i,[x_2]_i)$ are independent with every other coordinates in the input. In this case, we can decompose the multi-dimensional Gaussian kernel into products of one-dimensional Gaussian kernels in the following way
\begin{align}
	\textstyle{[A]_{x_1,x_2}\defeq P(x_1,x_2)/\sqrt{P(x_1)P(x_2)}=\prod_{i=1}^{d_1}P([x_1]_i,[x_2]_i)/\sqrt{P([x_1]_i)P([x_2]_i)}.}
\end{align}
Consequently, the singular values of $A$ will be the products of singular values of these one-dimensional Gaussian kernels:
$
	\{\prod_{i=1}^{d_1}|[\Sigma_{12}]_{i,i}|^{k_i}\}_{k_1,k_2,\cdots,k_{d_1}\ge 0}.
$
Therefore, the second largest singular value of $A$ will be $\max_{i\in[d_1]}|[\Sigma_{12}]_{i,i}|.$ Following the same reasoning as Theorem~\ref{thm:gaussian} we get 
$$\tau(P,Q,\calF)\le \frac{2}{1-\max_{i\in[d_1]}|[\Sigma_{12}]_{i,i}|}=\frac{2}{1-\sigma_{\rm max}(\Sigma_{12})}$$
where $\sigma_{\rm max}(\Sigma_{12})$ is the largest singular value of $\Sigma_{12}$. By noticing that $1-\sigma_{\rm max}(\Sigma_{12})=\lambda_{\rm min}(\Sigma_P)$ (Lemma~\ref{lem:cov-cor-eigenvalue}) we get Eq.~\eqref{equ:thm-gm-1}.

Now we turn to the general cases where $\Sigma_{12}$ is not diagonal. Recall that our model is $f(x)=f_1(x_1)+f_2(x_2)$ for some arbitrary functions $f_1,f_2$. Hence, we can rotate the inputs $x_1,x_2$ without affecting the model $\frer$. Formally speaking, for any orthogonal matrices $U\in\R^{d_1\times d_1}$ and $V\in\R^{d_2\times d_2}$, 
\begin{align}
	\sup_{f\in\calF}\frac{\E_Q[(f_1(x_1)+f_2(x_2))^2]}{\E_P[(f_1(x_1)+f_2(x_2))^2]}=\sup_{f\in\calF}\frac{\E_Q[(f_1(Ux_1)+f_2(Vx_2))^2]}{\E_P[(f_1(Ux_1)+f_2(Vx_2))^2]}.
\end{align}
If $U,V$ are the orthonormal matrices in the singular value decomposition $\Sigma_{12}=U^\top \Lambda_{12}V$ where $\Lambda_{12}$ is a diagonal matrix, we get
\begin{align}
	\E_{P}[(Ux_1) (Vx_2)^\top]=\Lambda_{12},\quad \E_{P}[(Ux_1)^\top (Ux_1)]=I,\quad \E_{P}[(Vx_2)^\top (Vx_2)]=I.
\end{align}
As a result, reusing the result from the previous case on inputs $Ux_1,Vx_2$ proves the desired result.

\paragraph{Remarks.} Our current techniques can only handle the case when the input is divided into $k=2$ subsets. This is because for $k\ge 3$ we must diagonalize multiple multi-dimensional Gaussian kernels simultaneously using the same set of eigenfunctions, as required in the proof of Theorem~\ref{thm:gaussian}. However, these multi-dimensional Gaussian kernels do not share the same eigenfunctions because the rotation matrix $U,V$ depends on the covariance $\E_P[x_ix_j^\top]$. Hence, the proof strategy for Theorem~\ref{thm:gaussian} fails for the case $k\ge 3$.

\section{Lower Bounds}\label{sec:lowerbounds}
In this section, we prove a lower bound as a motivation to consider structured distributions shifts. The following proposition shows that in the worst case, models learned on $P$ cannot extrapolate to $Q$ when the support of distribution $Q$ is not contained in the support of $P$.
\begin{proposition}\label{prop:lowerbound}
	Let the model class $\calF$ be the family of two-layer neural networks with ReLU activation:
	$
		\calF=\left\{\sum_{i}a_i\relu(w_i^\top x+b_i):w_i\in\R^{d},a_i,b_i\in\R\right\}.
	$
	Suppose for simplicity that all the inputs have unit norm (i.e., $\|x\|_2=1$).
	If $Q$ has non-zero probability mass on the set of points well-separated from the support of $P$ in the sense that 
	\begin{align}
		\exists \epsilon>0,\quad Q(\{x:\|x\|_2=1,\dist(x,\supp(P))\ge \epsilon\})>0,
	\end{align} we can construct a model $f\in\calF$ such that $\|f\|_P=0$ but $\|f\|_Q$ can be arbitrarily large.
\end{proposition}
A complete proof of this proposition is deferred to Appendix~\ref{app:pf-prop-lowerbound}. On a high level, we prove this proposition by construct a two-layer neural network $g_t$ that represents a bump function around any given input $t\in S^{d-1}$. As a result, when $t$ is a point in $\supp(Q)\setminus \supp(P)$, the model $g_t(x)$ will have zero $\ell_2$ norm on $P$ but have a positive $\ell_2$ norm on $Q$.
This construction is inspired by \citet[Theorem 5.1]{dong2021provable}. 

\section{Simulations with Synthetic Datasets}\label{sec:experiments}
In this section, we present the experiments that support our theory. The implementation details and ablation studies are deferred to Appendix~\ref{app:experiments}

We focus on the setting of Theorem~\ref{thm:gaussian-multi} where $x=(x_1,x_2)$ and $x_1\in \mathbb{R}^{d_1},x_2\in \mathbb{R}^{d_2}$, and $P,Q$ have matching marginals on $x_1,x_2$. We compare the extrapolation of the structured model $f_1(x_1)+f_2(x_2)$ (where $f_1,f_2$ are two-layer neural networks) and unstructured model $f(x)$ (where $f$ is a two-layer neural network with input $[x_1,x_2]\in\mathbb{R}^{d_1+d_2}$). The hidden dimension of the unstructured model $f(x)$ is two times larger than each component of the unstructured model, that is, $f_1,f_2$, hence the unstructured model is strictly more expressive than the structured model. The ground-truth label is generated by a random structured model $f^\star_1(x_1)+f^\star_2(x_2)$. Recall that our theory shows that the structured model provably extrapolates to the target distribution if $\lambda_{\min}(\Sigma_P)>0$ . Indeed, empirically we found that the structured model has a significantly smaller \emph{OOD loss} than the unstructured model even though the \emph{ID losses} are comparable, as shown in Table~\ref{tab:1}. Therefore, the structured model family indeed has a better extrapolation than the unstructured family.

\begin{table}[htp]
\begin{center}
	\begin{tabular}{ |c|c|c| } 
		\hline
		Model Class & ID loss & OOD loss \\ 
		\hline
		Structured model $f_1(x_1)+f_2(x_2)$ & 0.00100 & 0.00118 \\ 
		Unstructured model $f(x)$ & 0.00105	&  0.01235 \\ 
		\hline
	\end{tabular}
\end{center}
\caption{The ID and OOD losses of different model classes. The structured model has a significantly smaller \emph{OOD loss} than the unstructured model even though the \emph{ID losses} are comparable.}
\label{tab:1}
\end{table}
\section{Related Works}\label{sec:relatedworks}
The most related work is \citet{ben2010theory}, where they use the $\calH\Delta \calH$-distance to measure the maximum discrepancy of two models $f,g\in\calF$ on any distributions $P,Q$.  However, it remains an open question to determine when $\calH\Delta \calH$-distance is small for \emph{concrete} nonlinear model classes and distributions. On the technical side, the quantity $\tau$ is an analog of the $\calH\Delta \calH$-distance for regression problems, and we provide concrete examples where $\tau$ is upper bounded even if the distributions $P,Q$ have significantly different support.

Another closely related settings are domain adaptation \citep{ganin2015unsupervised,ghifary2016deep,ganin2016domain} and domain generalization \citep{gulrajani2020search,peters2016causal}, where the algorithm actively improve the extrapolation of learned model either by using unlabeled data from the test domain \citep{sun2016deep,li2020maximum,li2020enhanced,zhang2019bridging}, or learn an invariant model across different domains \citep{arjovsky2019invariant,peters2016causal,gulrajani2020search}.
In comparison, this paper studies a more basic question: whether a model trained on one distribution (without any implicit bias and unlabeled data from test domain) extrapolates to new distributions.
There are also prior works that theoretically analyze algorithms that use additional (unlabeled) data from the test distribution, such as self-training \citep{wei2020theoretical,chen2020self}, contrastive learning \citep{shen2022connect,haochen2022beyond}, label propagation \citep{cai2021theory}, etc.

\section{Conclusions}\label{sec:conclusions}
In this paper, we propose to study domain shifts between $P$ and $Q$ with the structure that each feature's marginal distribution has good overlap between source and target domain but the joint distribution of the features may have a much bigger shift.
As a first step toward understanding the extrapolation of nonlinear models, we prove sufficient conditions for the model $f(x)=\sum_{i=1}^{k}f_i(x_i)$ to extrapolate where $f_i$ is an arbitrary function of a single feature.
Even though the assumptions on the shift and function class is stylized, to the best of our knowledge, this is the first analysis of how concrete nonlinear models extrapolate when source and target distribution \emph{do not} have shared support.

There still remain many interesting open questions, which we leave as future works:
\begin{enumerate}
	\item Our current proof can only deal with a restricted nonlinear model family of the special form $f(x)=\sum_{i=1}^{k}f_i(x_i)$ due to technical reasons. Can we extend to a more general model class?
	\item In this paper, we focus on regression tasks with $\ell_2$ loss for mathematical simplicity, whereas majority of the prior works focus on the classification problems. Do similar results also hold for classification problem?
\end{enumerate}

\subsection*{Acknowledgment}
The authors would like to thank Yuanhao Wang, Yuhao Zhou, Hong Liu, Ananya Kumar, Jason D. Lee, and Kendrick Shen for helpful discussions. The authors would also like to thank the support from NSF CIF 2212263.

\bibliography{all.bib}
\bibliographystyle{plainnat}

\newpage
\appendix
\section*{List of Appendices}
\startcontents[sections]
\printcontents[sections]{l}{1}{\setcounter{tocdepth}{2}}
\newpage

\section{Missing Proofs}
In the following, we present the missing proofs.

\subsection{Proof of Proposition~\ref{prop:losses}}\label{app:pf-prop-losses}
In this section, we prove Proposition~\ref{prop:losses}.
\begin{proof}[Proof of Proposition~\ref{prop:losses}] By the definition of $\tau$, for any $f\in\calF$ we get
	\begin{align}
		\frac{\E_Q[(f(x)-f^\star(x))^2]}{\E_P[(f(x)-f^\star(x))^2]}\le \tau.
	\end{align}
	Or equivalently, 
	\begin{align}
		\E_Q[(f(x)-f^\star(x))^2]\le \tau\E_P[(f(x)-f^\star(x))^2].
	\end{align}
	As a result,
	\begin{align}
		&\E_{Q}[(y-f(x))^2]\le 2\E_{Q}[(y-f^\star(x))^2]+2\E_{Q}[(f(x)-f^\star(x))^2]\\
		\le\;& 4\erlzb+2\tau\E_{P}[(f(x)-f^\star(x))^2]\le 4\erlzb+4\tau\(\E_{P}[(y-f(x))^2]+\E_{P}[(y-f^\star(x))^2]\)\\
		\le\;& (8\tau+4)\erlzb+4\tau \E_{P}[(y-f(x))^2].
	\end{align}
\end{proof}

\subsection{Proof of Theorem~\ref{thm:discrete}}\label{app:pf-thm-discrete}
In this section, we prove Theorem~\ref{thm:discrete}.
\begin{proof}[Proof of Theorem~\ref{thm:discrete}]
	Let \begin{align}b_t(x_1,x_2)=\begin{cases}
			\ind{x_1=t},&\text{ when }t\le r_1,\\
			\ind{x_2=t-r_1},&\text{ when }r_1<t\le r_1+r_2.
	\end{cases}\end{align}
	\sloppy Then for any $f,f'\in\calF$, we can always find $v\in\R^{r_1+r_2}$ such that $f(x_1,x_2)-f'(x_1,x_2)=\sum_{i=1}^{r_1+r_2}v_ib_i(x_1,x_2)$ for all $x_1,x_2.$ Indeed, for any $f,f'\in\calF$ the architecture of our model implies $(f-f')(x_1,x_2)=(f_1-f_1')(x_1)+(f_2-f_2')(x_2).$ Therefore, we can simply set $v_t=(f_1-f_1')(t)$ for $1\le t\le r_1$ and $v_{t}=(f_2-f_2')(t-r_1)$ for $r_1<t\le r_2.$
	
	Consequently, 
	\begin{align}
		\tau\;&= \sup_{f,f'\in\calF}\frac{\E_Q[(f(x)-f'(x))^2]}{\E_P[(f(x)-f'(x))^2]}=\sup_{v\in\R^{r_1+r_2}}\frac{\E_Q[(\sum_{i=1}^{r_1+r_2}b_i(x_1,x_2)v_i)^2]}{\E_P[(\sum_{i=1}^{r_1+r_2}b_i(x_1,x_2)v_i)^2]}\\
		&=\sup_{v\in\R^{r_1+r_2}}\frac{\sum_{i,j=1}^{r_1+r_2}v_iv_j\E_Q[b_i(x_1,x_2)b_j(x_1,x_2)]}{\sum_{i,j=1}^{r_1+r_2}v_iv_j\E_P[b_i(x_1,x_2)b_j(x_1,x_2)]}.
	\end{align}
	The definition of $b_t(x_1,x_2)$ implies that for any distribution $P$, 
	\begin{align}
		\E_{P}[b_i(x_1,x_2)b_j(x_1,x_2)]=\begin{cases}
			\ind{i=j}P(x_1=i),&\text{when $1\le i,j\le r_1$},\\
			\ind{i=j}P(x_2=j),&\text{when $r_1<i,j\le r_1+r_2$},\\
			P(x_1=i,x_2=j),&\text{when $1\le i\le r_1<j\le r_1+r_2$},\\
			P(x_1=j,x_2=i),&\text{when $1\le j\le r_1<i\le r_1+r_2$}.
		\end{cases}
	\end{align}
	Then we have $[K_P]_{i,j}=\E_{P}[b_i(x_1,x_2)b_j(x_1,x_2)].$ Consequently, 
	\begin{align}
		\tau\;
		=\sup_{v\in\R^{r_1+r_2}}\frac{\sum_{i,j=1}^{r_1+r_2}v_iv_j\E_Q[b_j(x_1,x_2)b_j(x_1,x_2)]}{\sum_{i,j=1}^{r_1+r_2}v_iv_j\E_P[b_j(x_1,x_2)b_j(x_1,x_2)]}=\sup_{v\in\R^{r_1+r_2}}\frac{v^\top K_Qv}{v^\top K_P v},
	\end{align}
	
	Let $u\in\R^{r_1+r_2}$ such that $$[u]_i=\begin{cases}
		1,&\text{ when }i\le r_1,\\
		-1,&\text{ when }r_1<i\le r_2.
	\end{cases}$$
	We claim that $u$ is a eigenvector to both $K_Q$ and $K_P$ with eigenvalue 0. To see this, for any distribution $P$ and $i\in[r_1]$ we have
	\begin{align}
		&[K_P u]_i=P(x_1=i)[u]_i+\sum_{j=1}^{r_2} P(x_1=i,x_2=j)[u]_{r_1+j}\\
		=\;&P(x_1=i)-\sum_{j=1}^{r_2} P(x_1=i,x_2=j)=0.
	\end{align}
	Similarly for $i\in[r_2]$ we have $[K_P u]_{r_1+i}=0$. Combining these two cases together we prove $K_P u=0.$
	
	Then, by algebraic manipulation,
	\begin{align}
		&\sup_{v\in\R^{r_1+r_2}}\frac{v^\top K_Qv}{v^\top K_P v}=\sup_{v\in\R^{r_1+r_2},v\perp u}\frac{v^\top K_Qv}{v^\top K_P v}=\sup_{v\in\R^{r_1+r_2},v\perp u}\frac{v^\top D_Q^{1/2} \bar{K}_Q D_Q^{1/2}v}{v^\top D_P^{1/2} \bar{K}_P D_P^{1/2}v}\\
		\le\;&\frac{\lmax(\bar{K}_Q)}{\lambda_2(\bar{K}_P)}\sup_{v}\frac{\|D_Q^{1/2}v\|_2^2}{\|D_P^{1/2}v\|_2^2}\le \frac{\lmax(\bar{K}_Q)}{\lambda_2(\bar{K}_P)}\max_{i\in[2],t\in[r_i]}\frac{Q(x_i=t)}{P(x_i=t)}.
	\end{align}
	As a result, we only need to prove $\lmax(\bar{K}_Q)\le 2.$ To this end, note that
	\begin{align}
		&\lmax(\bar{K}_Q)=\sup_{w\in\R^{r_1+r_2}}\frac{w^\top \bar{K}_Q w}{w^\top w}=\sup_{v\in\R^{r_1+r_2}}\frac{v^\top K_Q v}{v^\top D_Q v}\\
		=\;&\sup_{v\in\R^{r_1+r_2}}\frac{\sum_{i\in[r_1],j\in[r_2]}P(x_1=i,x_2=j)([v]_i+[v]_{r_1+j})^2}{\sum_{i\in[r_1]}P(x_1=i)[v]_i^2+\sum_{j\in[r_2]}P(x_2=j)[v]_{r_1+j}^2}\\
		\le\;&\sup_{v\in\R^{r_1+r_2}}\frac{2\sum_{i\in[r_1],j\in[r_2]}P(x_1=i,x_2=j)([v]_i^2+[v]_{r_1+j}^2)}{\sum_{i\in[r_1]}P(x_1=i)[v]_i^2+\sum_{j\in[r_2]}P(x_2=j)[v]_{r_1+j}^2}\\
		\le\;&2.
	\end{align}
\end{proof}

\subsection{Extending Theorem~\ref{thm:discrete} to Multiple Dimensions ($k>2$).}\label{app:thm-discrete-multi}
In this section, we extend Theorem~\ref{thm:discrete} to the case where $x=(x_1,x_2,\cdots,x_k)$ with $k>2$. Following the same approach, we first define the (signless) Laplacian matrix $K_P.$

Without loss of generality, we assume that $x_i$ takes the value in $\{1,2, \cdots,r_i\}$ for all $i\in[k]$. Define $r=\sum_{i=1}^{k}r_i$. For a distribution $P$, define the matrix $P_{ij}\in\R^{r_i\times r_j}$ with entry $[P_{ij}]_{a,b}$ equals the probability mass $P(x_i=a, x_j=b)$ when $i\neq j$, and diagonal matrix $P_{ii}\in \R^{r_i\times r_i}$ with entry $[P_{ii}]_{a,a}$ equals the probability mass $P(x_i=a).$
In addition, let $K_P,\bar{K}_P\in\R^{r\times r}$ be the matrices defined as follows:
\begin{align}\label{equ:discrete-kernel-multi}
	K_P=\begin{pmatrix}
		P_{11} & P_{12} & \cdots & P_{1k}\\
		P_{21} & P_{22} & \cdots & P_{2k}\\
		\vdots & \vdots & \ddots & \vdots\\
		P_{k1} & P_{k2} & \cdots & P_{kk}
	\end{pmatrix},\quad \bar{K}_P=\diag(K_P)^{-1/2}K_P\diag(K_P)^{-1/2}.
\end{align}
Then the extrapolation power of the nonlinear model class $\calF$ is summarized in the following theorem.
\begin{theorem}\label{thm:discrete-multi}
	For any two dimensional distribution $P,Q$ over discrete random variables $x_1,x_2,\cdots,x_k$, we have 
	\begin{align}\label{equ:twovar-discrete-main-multi}
		\tau=\sup_{v\in\R^{r}}\frac{v^\top K_Qv}{v^\top K_P v}\le k\lambda_k(\bar{K}_P)^{-1}\max_{i\in[k],t\in[r_i]}\frac{Q(x_i=t)}{P(x_i=t)}.
	\end{align}
\end{theorem}
\begin{proof}[Proof of Theorem~\ref{thm:discrete-multi}]
	As discusses in the proof sketch, we first construct a set of basis for the model class $\calF=\{\sum_{i=1}^{k}f_i(x_i)\}.$ 
	For any $t\in[r]$, let $\iota(t)\in[k]$ be the index such that $\sum_{j=1}^{\iota(t)-1}r_j<t\le \sum_{j=1}^{\iota(t)}r_j$ and define $\tau(t)=t-\sum_{j=1}^{\iota(t)-1}r_j.$ Then we define
	\begin{align}
		b_t(x)=\ind{x_{\iota(t)}=\tau(t)}.
	\end{align}
	
	Then for any $f\in\calF$, we can always find $v\in\R^{r}$ such that $f(x)=\sum_{i=1}^{r}[v]_ib_i(x)$ for all $x\in\calX.$ Indeed, for any $f\in\calF$ the architecture of our model implies $f(x)=\sum_{i=1}^{k}f_i(x_i).$ Therefore, we can simply set $v_t=f_{\iota(t)}\(\tau(t)\)$ for $t\in[r]$.
	
	Since $\calF$ is closed under addition, we get
	\begin{align}
		\tau\;&= \sup_{f,f'\in\calF}\frac{\E_Q[(f(x)-f'(x))^2]}{\E_P[(f(x)-f'(x))^2]}= \sup_{f\in\calF}\frac{\E_Q[(f(x))^2]}{\E_P[(f(x))^2]}\\
		&=\sup_{v\in\R^{r}}\frac{\E_Q[(\sum_{i=1}^{r}b_i(x)[v]_i)^2]}{\E_P[(\sum_{i=1}^{r}b_i(x)[v]_i)^2]}\\
		&=\sup_{v\in\R^{r}}\frac{\sum_{i,j=1}^{r}[v]_i[v]_j\E_Q[b_i(x)b_j(x)]}{\sum_{i,j=1}^{r}[v]_i[v]_j\E_P[b_i(x)b_j(x)]}.
	\end{align}
	The definition of $b_t(x)$ implies that for any distribution $P$, 
	\begin{align}
		\E_{P}[b_i(x)b_j(x)]=P(x_{\iota(i)}=\tau(i),x_{\iota(j)}=\tau(j)).
	\end{align}
	Then the kernel matrix is given by $[K_P]_{i,j}=\E_{P}[b_i(x)b_j(x)],$ which is exactly the definition in Eq.~\eqref{equ:discrete-kernel-multi}. Consequently, 
	\begin{align}
		\tau\;
		=\sup_{v\in\R^{r}}\frac{\sum_{i,j=1}^{r}[v]_i[v]_j\E_Q[b_i(x)b_j(x)]}{\sum_{i,j=1}^{r}[v]_i[v]_j\E_P[b_i(x)b_j(x)]}=\sup_{v\in\R^{r}}\frac{v^\top K_Qv}{v^\top K_P v},
	\end{align}
	which proves the first part of Eq.~\eqref{equ:twovar-discrete-main-multi}. 
	
	Now we prove the second part the theorem. To this end, we first characterize the shared eigenvectors of $K_P$ and $K_Q$.
	
	For any $1\le t\le r-1,$ define the vector $u_t\in \R^{r}$ where $$[u_1]_i=\begin{cases}
		1,&\text{ when }\iota(i)=t,\\
		-1,&\text{ when }\iota(i)=t+1\\
		0,&\text{ o.w. }
	\end{cases}$$
	We claim that $u_t$ is a eigenvector to both $K_Q$ and $K_P$ with eigenvalue 0. To see this, for any distribution $P$ and $i\in[r]$ we have
	\begin{align}
		&[K_P u_t]_i=\sum_{j:\iota(j)=t}P(x_{\iota(i)}=\tau(i),x_t=\tau(j))-\sum_{j:\iota(j)=t+1}P(x_{\iota(i)}=\tau(i),x_{t+1}=\tau(j))\\
		=\;&P(x_{\iota(i)}=\tau(i))-P(x_{\iota(i)}=\tau(i))=0.
	\end{align}
	Let $U={\rm span}(u_1,\cdots,u_{r-1})$ be the linear subspace of such eigenvectors. It follows that 
	\begin{align}
		u^\top K_P u=u^\top K_Q u=0,\quad\forall u\in U.
	\end{align} 
	Because $U$ is the linear subspace of eigenvectors of $K_P$ (and $K_Q$) corresponding to eigenvalues 0, which is the minimum eigenvalue of $K_P$ (and $K_Q$), we get
	\begin{align}
		&\sup_{v\in\R^{r}}\frac{v^\top K_Qv}{v^\top K_P v}=\sup_{v\in\R^{r},v\perp U}\frac{v^\top K_Qv}{v^\top K_P v}=\sup_{v\in\R^{r},v\perp U}\frac{v^\top D_Q^{1/2} \bar{K}_Q D_Q^{1/2}v}{v^\top D_P^{1/2} \bar{K}_P D_P^{1/2}v}\\
		\le\;&\frac{\lmax(\bar{K}_Q)}{\lambda_k(\bar{K}_P)}\sup_{v}\frac{\|D_Q^{1/2}v\|_2^2}{\|D_P^{1/2}v\|_2^2}\le 	\frac{\lmax(\bar{K}_Q)}{\lambda_k(\bar{K}_P)}\max_{i\in[k],t\in[r_i]}\frac{Q(x_i=t)}{P(x_i=t)}.
	\end{align}
	As a result, we only need to prove $\lmax(\bar{K}_Q)\le k.$ To this end, for any $i\in[k]$, let $s_i=\sum_{j=1}^{i-1}r_j.$ Note that
	\begin{align}
		&\lmax(\bar{K}_Q)=\sup_{u\in\R^{r}}\frac{u^\top \bar{K}_Q u}{u^\top u}=\sup_{v\in\R^{r}}\frac{v^\top K_Q v}{v^\top D_Q v}\\
		=\;&\sup_{v\in\R^{r}}\frac{\sum_{j_1\in[r_1],j_2\in[r_2],\cdots,j_k\in[r_k]}P(x_1=j_1,x_2=j_2,\dots,x_k=j_k)\(\sum_{i=1}^{k}[v]_{s_i+j_i}\ind{x_i=j_i}\)^2}{\sum_{i\in[k],j\in[r_k]}P(x_i=j)[v]_{s_i+j}^2}\\
		\le\;&\sup_{v\in\R^{r}}\frac{k\sum_{j_1\in[r_1],j_2\in[r_2],\cdots,j_k\in[r_k]}P(x_1=j_1,x_2=j_2,\dots,x_k=j_k)\(\sum_{i=1}^{k}\ind{x_i=j_i}[v]_{s_i+j_i}^2\)}{\sum_{i\in[k],j\in[r_k]}P(x_i=j)[v]_{s_i+j}^2}\\
		=\;&\sup_{v\in\R^{r}}\frac{k\sum_{i\in[k],j\in[r_k]}P(x_i=j)[v]_{s_i+j}^2}{\sum_{i\in[k],j\in[r_k]}P(x_i=j)[v]_{s_i+j}^2}\\
		\le\;&k.
	\end{align}
\end{proof}

\subsection{Proof of Theorem~\ref{thm:gaussian}}\label{app:pf-thm-gaussian}
In this section, we prove Theorem~\ref{thm:gaussian}.
\begin{proof}[Proof of Theorem~\ref{thm:gaussian}]
	Now consider any fixed model pairs $f,f'\in\calF$. Let $g_i(x_i)=f_i(x_i)-f_i'(x_i)$ and $g=\sum_{i=1}^{d}g_i(x_i).$ First we assume the marginals satisfies $x_i\sim \calN(0,1)$ for every $i\in[d]$. As a result,
	Lemma~\ref{lem:orthonormal-basis} implies that there exists coefficients $\{\alpha_i^{(n)}\}_{i\in[d],n\in\Z_+}$ with
	\begin{align}
		g_i(x_i)\sqrt{P(x_i)}=\sum_{n\ge 0}\alpha_i^{(n)}\psi_n(x_i),\quad\forall i\in[d],
	\end{align}
	where $\{\psi_n(\cdot)\}_{n\in\Z_+}$ are a set of orthonormal basis of $L^2(\R)$ defined by $$\psi_n(x)\defeq H_n\(\frac{1}{\sqrt{2}}x\)\exp\(-\frac{1}{4}x^2\)(2\pi)^{-1/4}(2^n n!)^{-1/2}.$$ As a result,
	\begin{align}
		&\E_P\[\(\sum_{i=1}^{d}g_i(x_i)\)^2\]=\int \sum_{i,j=1}^{d}g_i(x_i)g_j(x_j)P(x_1,\cdots,x_d)\dd x_1\cdots x_d\\
		=\;&\sum_{i=1}^{d}\int g_i(x_i)^2P(x_i)\dd x_i+2\sum_{1\le i<j\le d}\int g_i(x_i)g_j(x_j)P(x_i,x_j)\dd x_ix_j\\
		=\;&\sum_{i=1}^{d}\int \Big(\sum_{n\ge 0}\alpha_i^{(n)}\psi_n(x_i)\Big)^2\dd x_i\\
		&\quad+2\sum_{1\le i<j\le d}\int (g_i(x_i)\sqrt{P(x_i)})(g_j(x_j)\sqrt{P(x_j)})\frac{P(x_i,x_j)}{\sqrt{P(x_i)P(x_j)}}\dd x_ix_j\\
		=\;&\sum_{i=1}^{d}\sum_{n\ge 0}(\alpha_i^{(n)})^2
		+2\sum_{1\le i<j\le d}\int \Big(\sum_{n\ge 0}\alpha_i^{(n)}\psi_n(x_i)\Big)\Big(\sum_{n\ge 0}\alpha_j^{(n)}\psi_n(x_j)\Big)\frac{P(x_i,x_j)}{\sqrt{P(x_i)P(x_j)}}\dd x_ix_j.\label{equ:pf-scalar-1}
	\end{align}
	By Theorem~\ref{thm:gaussian-kernel}, we have
	\begin{align}
		\frac{P(x_i,x_j)}{\sqrt{P(x_i)P(x_j)}}=\sum_{n\ge 0}([\Sigma_P]_{i,j})^n\psi_n(x_i)\psi_n(x_j).
	\end{align}
	Continuing the second term of Eq.~\eqref{equ:pf-scalar-1} we get
	\begin{align}
		&\sum_{1\le i<j\le d}\int \Big(\sum_{n\ge 0}\alpha_i^{(n)}\psi_n(x_i)\Big)\Big(\sum_{n\ge 0}\alpha_j^{(n)}\psi_n(x_i)\Big)\frac{P(x_i,x_j)}{\sqrt{P(x_i)P(x_j)}}\dd x_ix_j\\
		=\;&\sum_{1\le i<j\le d}\int \Big(\sum_{n\ge 0}\alpha_i^{(n)}\psi_n(x_i)\Big)\Big(\sum_{n\ge 0}\alpha_j^{(n)}\psi_n(x_i)\Big)\Big(\sum_{n\ge 0}([\Sigma_P]_{i,j})^n\psi_n(x_i)\psi_n(x_j)\Big)\dd x_ix_j\\
		=\;&\sum_{1\le i<j\le d} \sum_{n\ge 0}\alpha_i^{(n)}\alpha_j^{(n)}([\Sigma_P]_{i,j})^n.
	\end{align}
	As a result, 
	\begin{align}
		\E_P\[\(\sum_{i=1}^{d}g_i(x_i)\)^2\]=\sum_{i=1}^{d}\sum_{n\ge 0}(\alpha_i^{(n)})^2+2\sum_{1\le i<j\le d} \sum_{n\ge 0}\alpha_i^{(n)}\alpha_j^{(n)}([\Sigma_P]_{i,j})^n.
	\end{align}
	Define $\kappa=\sup_{n\ge 0}\sup_{v\in\R^{d}}\frac{v^\top (\Sigma_{\rm Q})^{\odot n} v}{v^\top (\Sigma_{\rm P})^{\odot n} v}.$ In the following, we prove \begin{align}\label{equ:pf-scalar-2}
		\E_Q\[\(\sum_{i=1}^{d}g_i(x_i)\)^2\]\le \kappa\E_P\[\(\sum_{i=1}^{d}g_i(x_i)\)^2\].
	\end{align}
	For any fixed $n\ge 0$, let $\Sigma_P^{(n)}\in\R^{d\times d}$ be the matrix where $[\Sigma_P^{(n)}]_{i,j}=([\Sigma_P]_{i,j})^n$ (define $\Sigma_Q^{(n)}$ similarly), and $\vec{\alpha}^{(n)}\in\R^{d}$ the vector where $[\vec{\alpha}^{(n)}]_i=\alpha_i^{(n)}.$ Then 
	\begin{align*}
		\E_P\[\(\sum_{i=1}^{d}g_i(x_i)\)^2\]=\sum_{n\ge 0}(\vec{\alpha}^{(n)})^\top \Sigma_P^{(n)}\vec{\alpha}^{(n)},\quad \E_Q\[\(\sum_{i=1}^{d}g_i(x_i)\)^2\]=\sum_{n\ge 0}(\vec{\alpha}^{(n)})^\top \Sigma_Q^{(n)}\vec{\alpha}^{(n)}.
	\end{align*}
	By the definition of $\kappa$ we get $\Sigma_Q^{(n)}\preceq \kappa\Sigma_P^{(n)}$ for all $n\ge 0$. Consequently,
	\begin{align}
		\sum_{n\ge 0}(\vec{\alpha}^{(n)})^\top \Sigma_Q^{(n)}\vec{\alpha}^{(n)}\le \kappa \Big(\sum_{n\ge 0}(\vec{\alpha}^{(n)})^\top \Sigma_P^{(n)}\vec{\alpha}^{(n)}\Big),
	\end{align}
	which implies Eq.~\eqref{equ:pf-scalar-2}.
	
	Now we prove $\kappa\le \frac{d}{\lmin(\Sigma_P)}.$ Since $\Sigma_Q$ is a covariance matrix with $\diag(\Sigma_Q)=I$, we have $\|\Sigma_Q^{\odot n}\|_2\le d\|\Sigma_Q^{\odot n}\|_\infty\le d.$ In addition, Lemma~\ref{lem:elementwise-product-lmin} proves that $\lmin(\Sigma_P^{\odot n})\ge \lmin(\Sigma_P)$ for every $n\ge 1.$ As a result,
	$$\inf_{n\ge 1}\inf_{v\in\R^{d}}(v^\top (\Sigma_{\rm P})^{\odot n} v)^{-1}\ge \inf_{n\ge 1}\lmin(\Sigma_P^{\odot n})\ge \lmin(\Sigma_P).$$
	Since $n=0\implies (\Sigma_{\rm P})^{\odot n}=(\Sigma_{\rm Q})^{\odot n}$, we get
	\begin{align}
		\kappa=\sup_{n\ge 0}\sup_{v\in\R^{d}}\frac{v^\top (\Sigma_{\rm Q})^{\odot n} v}{v^\top (\Sigma_{\rm P})^{\odot n} v}=\max\left\{1,\sup_{n\ge 1}\sup_{v\in\R^{d}}\frac{v^\top (\Sigma_{\rm Q})^{\odot n} v}{v^\top (\Sigma_{\rm P})^{\odot n} v}\right\}\le \max\left\{1,\frac{d}{\lmin(\Sigma_P)}\right\}.
	\end{align}
	By the fact that $\lmin(\Sigma_P)\le d\|\Sigma_P\|_\infty\le d$, we prove the desired result for the special case where $x_i\sim \calN(0,1)$ for every $i\in[d]$.
	
	For the general case where $x_i$ has arbitrary mean and variance, Recall that $\tilde{x}=(\tilde{x}_1,\cdots,\tilde{x}_d)$ is the normalized input where $\tilde{x}_i\defeq (x_i-\E_P[x_i])\Var(x_i)^{-1/2}$ has zero mean and unit variance for every $i\in[d]$, and $\tilde{\Sigma}_P\defeq \E_P[\tilde{x}\tilde{x}^\top]$ is the covariance matrix of $\tilde{x}$. Let $\tilde{P},\tilde{Q}$ be the distribution of random variable $\tilde{x}$ on the source/target distribution $P,Q$, respectively. Applying the same argument to features $\tilde{x}$, we get $\tau(\tilde{P},\tilde{Q},\calF)\le \frac{d}{\lmin(\tilde{\Sigma}_P)}.$ Combining with Lemma~\ref{lem:normalize}, which proves that normalization does not change the $\frer$, we get 
	\begin{align}
		\tau(P,Q,\calF)=\tau(\tilde{P},\tilde{Q},\calF)\le \frac{d}{\lmin(\tilde{\Sigma}_P)}.
	\end{align}
\end{proof}

In the following, we state and proof Lemma~\ref{lem:normalize}.
\begin{lemma}\label{lem:normalize}
	For any distributions $P,Q$ with matching marginals, define $\tilde{P}$ to be the density of the random variable $\tilde{x}=(\tilde{x}_1,\cdots,\tilde{x}_d)$ where $\tilde{x}_i\defeq\Var(x_i)^{-1/2}(x_i-\E_P[x_i])$, $x\sim P$ (and define $\tilde{Q}$ similarly). Then we have
	\begin{align}
		\tau(P,Q,\calF)=\tau(\tilde{P},\tilde{Q},\calF).
	\end{align}
\end{lemma}
\begin{proof}
	Let $\{h_i:\R\to\R\}_{i=1,2,\cdots,d}$ be any set of invertible functions. Since for every $f=\sum_{i}f_i(x_i)\in\calF$ there exists $f\circ h\defeq \sum_{i}f_i(h_i(x_i))\in\calF$ and vise versa, we get
	\begin{align}
		\sup_{f\in\calF}\frac{\E_Q[\sum_i f_i(x_i)^2]}{\E_P[\sum_i f_i(x_i)^2]}=\sup_{f\in\calF}\frac{\E_Q[\sum_i f_i(h_i(x_i))^2]}{\E_P[\sum_i f_i(h_i(x_i))^2]}.
	\end{align}
	Let $\tilde{P}$ be the density of the random variable $(h_1(x_1),\cdots,h_d(x_d))$ when $x\sim P$ (and define $\tilde{Q}$ similarly). Then we have
	\begin{align}
		\frac{\E_{x\sim Q}[\sum_i f_i(h_i(x_i))^2]}{\E_{x\sim P}[\sum_i f_i(h_i(x_i))^2]}=\frac{\E_{t\sim \tilde{Q}}[\sum_i f_i(t_i)^2]}{\E_{t\sim \tilde{P}}[\sum_i f_i(t_i)^2]}.
	\end{align}
	Consequently, 
	\begin{align}
		&\tau(P,Q,\calF)=\sup_{f\in\calF}\frac{\E_Q[\sum_i f_i(x_i)^2]}{\E_P[\sum_i f_i(x_i)^2]}=\sup_{f\in\calF}\frac{\E_Q[\sum_i f_i(h_i(x_i))^2]}{\E_P[\sum_i f_i(h_i(x_i))^2]}\\
		=\;&\sup_{f\in\calF}\frac{\E_{t\sim \tilde{Q}}[\sum_i f_i(t_i)^2]}{\E_{t\sim \tilde{P}}[\sum_i f_i(t_i)^2]}=\tau(\tilde{P},\tilde{Q},\calF).
	\end{align}
	
	Finally, this lemma is proved by taking $h_i(x_i)=\Var(x_i)^{-1/2}(x_i-\E_P[x_i])$.
\end{proof}

\subsection{Proof of Theorem~\ref{thm:gaussian-multi}}\label{app:pf-thm-gaussian-multi}
In this section, we prove Theorem~\ref{thm:gaussian-multi}.
\begin{proof}[Proof of Theorem~\ref{thm:gaussian-multi}]
	Without loss of generality, we assume $d_2\le d_1.$ Since $\calF$ is closed under addition, we have $$\tau= \max_{g\in\calF}\frac{\E_Q[(g_1(x_1)+g_2(x_2))^2]}{\E_P[(g_1(x_1)+g_2(x_2))^2]}.$$
	In the following, we first prove \begin{align}\label{equ:pf-thm-gm-1}
		\tau\le \frac{2}{1-\sigma_{\rm max}(\Sigma_{12})}
	\end{align} by considering two cases separately.
	
	\paragraph{Case 1: when $g=(g_1,g_2)\in\calF$ satisfies $\E_P[g_1(x_1)]=\E_P[g_2(x_2)]=0$.}  Since $P$ and $Q$ have matching marginals on $x_1$ and $x_2$ respectively, we get
	\begin{align}
		\E_Q[g_1(x_1)^2+g_2(x_2)^2]=&\E_Q[g_1(x_1)^2]+\E_Q[g_2(x_2)^2]\\
		=\;&\E_P[g_1(x_1)^2]+\E_P[g_2(x_2)^2]=\E_P[(g_1(x_1)+g_2(x_2))^2].
	\end{align}
	As a result, by the definition of $\tau$ we have
	\begin{align*}
		&\frac{\E_Q[(g_1(x_1)+g_2(x_2))^2]}{\E_P[(g_1(x_1)+g_2(x_2))^2]}
		=\frac{\E_Q[(g_1(x_1)+g_2(x_2))^2]}{\E_Q[g_1(x_1)^2+g_2(x_2)^2]}\frac{\E_P[g_1(x_1)^2+g_2(x_2)^2]}{\E_P[(g_1(x_1)+g_2(x_2))^2]}\\
		\le\;&\frac{\E_Q[2(g_1(x_1)^2+g_2(x_2)^2)]}{\E_Q[g_1(x_1)^2+g_2(x_2)^2]}\frac{\E_P[g_1(x_1)^2+g_2(x_2)^2]}{\E_P[(g_1(x_1)+g_2(x_2))^2]}= 2\frac{\E_P[g_1(x_1)^2+g_2(x_2)^2]}{\E_P[(g_1(x_1)+g_2(x_2))^2]}.
	\end{align*}
	Hence, we only need to prove 
	\begin{align}\label{equ:pf-multi-3}
		(1-\sigma_{\rm max}(\Sigma_{12}))\E_P[g_1(x_1)^2+g_2(x_2)^2]\le \E_P[(g_1(x_1)+g_2(x_2))^2].
	\end{align}
	
	Let $\Sigma_{12}=U\Lambda V^\top$ be the singular value decomposition of $\Sigma_{12}$, where $\Lambda\in\R^{d_1\times d_2}$ is a diagonal matrix with entries $\sigma_1,\cdots,\sigma_{d_2}.$  
	
	Following Theorem~\ref{thm:gaussian-kernel-multi}, we define $t_1=U^\top x_1\in\R^{d_1}, t_2=V^\top x_2\in\R^{d_2}$. Consequently, $\E_P[t_1t_2^\top]=\Lambda$ and 
	\begin{align}
		(t_1,t_2)\sim \calN\(0,\begin{pmatrix}
			I_{d_1} & \Lambda\\ \Lambda^\top & I_{d_2}
		\end{pmatrix}\).
	\end{align}
	Let $P'(t_1,t_2)$ be the density of the distribution of $(t_1,t_2)$, and $P'(t_1),P'(t_2)$ the density of corresponding marginals. As a result,
	\begin{align}
		&\E_{P}[(g_1(x_1)+g_2(x_2))^2]=\E_{P'}[(g_1(t_1)+g_2(t_2))^2]\\
		=\;&\sum_{i=1}^{2}\int g_i(t_i)^2P'(t_i)\dd t_i+2\int g_1(t_1)g_2(t_2)P'(t_1,t_2)\dd t_1t_2.\label{equ:pf-multi-1}
	\end{align}
	Since $g_i(t_i)\sqrt{P'(t_i)}$ is square integrable, there exists weights $\{\alpha_i^{(n_1,\cdots,n_{d_i})}\}_{n_1,\cdots,n_{d_i}\ge 0}$ such that
	\begin{align}
		\forall i\in[2],\quad g_i(t_i)\sqrt{P'(t_i)}=\sum_{n_1,\cdots,n_{d_i}\ge 0} \alpha_i^{(n_1,\cdots,n_{d_i})}\prod_{j=1}^{d_i}\psi_{n_j}([t_i]_j),\label{equ:pf-multi-2}
	\end{align}
	where $\{\prod_{j=1}^{d_i}\psi_{n_j}([t_i]_j)\}_{n_1,\cdots,n_{d_i}\ge 0}$ forms an orthonormal basis of $L^2(\R^{d_i}).$
	Consequently, 
	\begin{align}
		&\E_P[g_1(x_1)^2+g_2(x_2)^2]=\E_{P'}[g_1(t_1)^2+g_2(t_2)^2]\\
		=\;&\sum_{n_1,\cdots,n_{d_1}}\(\alpha_1^{(n_1,\cdots,n_{d_1})}\)^2+\sum_{n_1,\cdots,n_{d_2}}\(\alpha_2^{(n_1,\cdots,n_{d_2})}\)^2.
	\end{align}
	
	Now we turn to the RHS of Eq.~\eqref{equ:pf-multi-3}. Continuing Eq.~\eqref{equ:pf-multi-1} and apply Theorem~\ref{thm:gaussian-kernel-multi} we get,
	\begin{align*}
		&\sum_{i=1}^{2}\int g_i(t_i)^2P'(t_i)\dd t_i		=\sum_{i=1}^{2}\sum_{n_1,\cdots,n_{d_i}}\(\alpha_i^{(n_1,\cdots,n_{d_i})}\)^2
	\end{align*} and \begin{align*}
		&\int g_1(t_1)g_2(t_2)P'(t_1,t_2)\dd t_1t_2\\
		=\;&\int\prod_{i=1}^{2}\(\sum_{n_1,\cdots,n_{d_1}\ge 0} \alpha_i^{(n_1,\cdots,n_{d_i})}\prod_{j=1}^{d_i}\psi_{n_j}([t_i]_j)\)\prod_{i=1}^{d_2}\(\sum_{n=0}^{\infty}\sigma_i^n \psi_n([t_1]_i)\psi_n([t_2]_i)\)\prod_{i=d_2+1}^{d_1}\psi_0([t_1]_i)\dd t_1t_2\\
		=\;&\sum_{n_1,\cdots,n_{d_2}\ge 0}\alpha_1^{(n_1,\cdots,n_{d_2},0,\cdots,0)}\alpha_2^{(n_1,\cdots,n_{d_2})}\prod_{i=1}^{d_2}\sigma_i^{n_i}.
	\end{align*}
	As a result, \begin{align}
		&\E_{P'}[(g_1(t_1)+g_2(t_2))^2]\\
		=\;&\sum_{i=1}^{2}\sum_{n_1,\cdots,n_{d_i}}\(\alpha_i^{(n_1,\cdots,n_{d_i})}\)^2+2\sum_{n_1,\cdots,n_{d_2}\ge 0}\alpha_1^{(n_1,\cdots,n_{d_2},0,\cdots,0)}\alpha_2^{(n_1,\cdots,n_{d_2})}\prod_{i=1}^{d_2}\sigma_i^{n_i}\\
		=\;&\sum_{n_1,\cdots,n_{d_2}}\begin{pmatrix}
			\alpha_1^{(n_1,\cdots,n_{d_2},0,\cdots,0)}\\\alpha_2^{(n_1,\cdots,n_{d_2})}
		\end{pmatrix}^\top\begin{pmatrix}
			1 & \prod_{i=1}^{d_2}\sigma_i^{n_i}\\
			\prod_{i=1}^{d_2}\sigma_i^{n_i} & 1
		\end{pmatrix}\begin{pmatrix}
			\alpha_1^{(n_1,\cdots,n_{d_2},0,\cdots,0)}\\\alpha_2^{(n_1,\cdots,n_{d_2})}
		\end{pmatrix}\notag\\
		&\quad + \sum_{n_1,\cdots,n_{d_1}}\ind{n_{d_2+1}+\cdots+n_{d_1}\neq 0}\(\alpha_i^{(n_1,\cdots,n_{d_i})}\)^2.\label{equ:pf-multi-4}
	\end{align}
	Now for any $n_1,\cdots,n_{d_2}\ge 0$, consider the matrix $$\begin{pmatrix}
		1 & \prod_{i=1}^{d_2}\sigma_i^{n_i}\\
		\prod_{i=1}^{d_2}\sigma_i^{n_i} & 1
	\end{pmatrix}.$$
	When $n_1+\cdots+n_{d_2}\neq 0$, we have $\prod_{i=1}^{d_2}\sigma_i^{n_i}\le \max_{i\in[d_2]}\sigma_i$ because Lemma~\ref{lem:upperbound-singularvalue} implies $\sigma_i\le 1$ for every $i\in[d_2]$. Therefore in this case,
	\begin{align}
		&\begin{pmatrix}
			\alpha_1^{(n_1,\cdots,n_{d_2},0,\cdots,0)}\\\alpha_2^{(n_1,\cdots,n_{d_2})}
		\end{pmatrix}^\top\begin{pmatrix}
			1 & \prod_{i=1}^{d_2}\sigma_i^{n_i}\\
			\prod_{i=1}^{d_2}\sigma_i^{n_i} & 1
		\end{pmatrix}\begin{pmatrix}
			\alpha_1^{(n_1,\cdots,n_{d_2},0,\cdots,0)}\\\alpha_2^{(n_1,\cdots,n_{d_2})}
		\end{pmatrix}\\
		\ge&\; (1-\sigma_{\rm max}(\Sigma_{12}))\(\(\alpha_1^{(n_1,\cdots,n_{d_2},0,\cdots,0)}\)^2+\(\alpha_2^{(n_1,\cdots,n_{d_2})}\)^2\).
	\end{align}
	When $n_1+\cdots+n_{d_2}=0$, by Eq.~\eqref{equ:pf-multi-2} we get
	\begin{align}
		&\alpha_1^{(0,\cdots,0)}=\int g_1(t_1)\sqrt{P'(t_1)}\prod_{j=1}^{d_1}\psi_{0}([t_1]_j)\dd t_1\\
		=\;&\int g_1(t_1)\sqrt{P'(t_1)}\prod_{j=1}^{d_i}\sqrt{P'([t_1]_j)}\dd t_1\\
		=\;&\int g_1(t_1)P'(t_1)\dd t_1=\E_{P'}[g_1(t_1)]=\E_{P}[g_1(x_1)]=0.
	\end{align}
	Similarly, $\alpha_2^{(0,\cdots,0)}=0.$ As a result, when $n_1+\cdots+n_{d_2}=0$ we have
	\begin{align}
		&\begin{pmatrix}
			\alpha_1^{(n_1,\cdots,n_{d_2},0,\cdots,0)}\\\alpha_2^{(n_1,\cdots,n_{d_2})}
		\end{pmatrix}^\top\begin{pmatrix}
			1 & \prod_{i=1}^{d_2}\sigma_i^{n_i}\\
			\prod_{i=1}^{d_2}\sigma_i^{n_i} & 1
		\end{pmatrix}\begin{pmatrix}
			\alpha_1^{(n_1,\cdots,n_{d_2},0,\cdots,0)}\\\alpha_2^{(n_1,\cdots,n_{d_2})}
		\end{pmatrix}\\
		=\;&\(\(\alpha_1^{(n_1,\cdots,n_{d_2},0,\cdots,0)}\)^2+\(\alpha_2^{(n_1,\cdots,n_{d_2})}\)^2\)=0.
	\end{align}
	Combining these two cases together, we get
	\begin{align}
		&\sum_{n_1,\cdots,n_{d_2}}\begin{pmatrix}
			\alpha_1^{(n_1,\cdots,n_{d_2},0,\cdots,0)}\\\alpha_2^{(n_1,\cdots,n_{d_2})}
		\end{pmatrix}^\top\begin{pmatrix}
			1 & \prod_{i=1}^{d_2}\sigma_i^{n_i}\\
			\prod_{i=1}^{d_2}\sigma_i^{n_i} & 1
		\end{pmatrix}\begin{pmatrix}
			\alpha_1^{(n_1,\cdots,n_{d_2},0,\cdots,0)}\\\alpha_2^{(n_1,\cdots,n_{d_2})}
		\end{pmatrix}\\
		\ge \;&(1-\sigma_{\rm max}(\Sigma_{12}))\(\sum_{n_1,\cdots,n_{d_2}}\(\alpha_1^{(n_1,\cdots,n_{d_2},0,\cdots,0)}\)^2+\(\alpha_2^{(n_1,\cdots,n_{d_2})}\)^2\).
	\end{align}
	Plug in to Eq.~\eqref{equ:pf-multi-4} we get
	\begin{align}
		&\E_{P}[(g_1(x_1)+g_2(x_2))^2]=\E_{P'}[(g_1(t_1)+g_2(t_2))^2]\\
		\ge \;&(1-\sigma_{\rm max}(\Sigma_{12}))\(\sum_{n_1,\cdots,n_{d_1}}\(\alpha_1^{(n_1,\cdots,n_{d_1})}\)^2+\sum_{n_1,\cdots,n_{d_2}}\(\alpha_2^{(n_1,\cdots,n_{d_2})}\)^2\)\\
		=\;&(1-\sigma_{\rm max}(\Sigma_{12}))\E_P[g_1(x_1)^2+g_2(x_2)^2],
	\end{align}
	which finished the proof for Eq.~\eqref{equ:pf-multi-3}.
	
	\paragraph{Case 2: when $g=(g_1,g_2)\in\calF$ satisfies $\E_P[g_1(x_1)+g_2(x_2)]=0$.} Let $\mu=\E_P[g_1(x_1)].$ Define $g'=(g_1',g_2')$ where $g_1'(x_1)=g_1(x_1)-\mu,g_2'(x_2)=g_2(x_2)+\mu.$ By the definition of $\calF$ we get $g'\in\calF$, and $g'$ satisfies $\E_P[g_1(x_1)]=\E_P[g_2(x_2)]=0$. Because $g_1(x_1)+g_2(x_2)=g_1'(x_1)+g_2'(x_2)$ for all $x_1\in\R^{d_1},x_2\in\R^{+d_2}$, plugging in the result of case 1 we get
	\begin{align}
		\frac{\E_Q[(g_1(x_1)+g_2(x_2))^2]}{\E_P[(g_1(x_1)+g_2(x_2))^2]}=\frac{\E_Q[(g'_1(x_1)+g'_2(x_2))^2]}{\E_P[(g'_1(x_1)+g'_2(x_2))^2]}\le \frac{1}{1-\sigma_{\rm max}(\Sigma_{12})}.
	\end{align}
	
	\paragraph{Case 3: when $g=(g_1,g_2)\in\calF$.} Now we consider the most general case. Let $\nu=\E_P[g_1(x_1)+g_2(x_2)].$ Since $P$ and $Q$ have matching marginals on $x_1$ and $x_2$ respectively, we get
	\begin{align}
		\E_Q[g_1(x_1)+g_2(x_2)]=&\E_Q[g_1(x_1)]+\E_Q[g_2(x_2)]\\
		=\;&\E_P[g_1(x_1)]+\E_P[g_2(x_2)]=\E_P[g_1(x_1)+g_2(x_2)]=\nu.
	\end{align}
	Define $g'=(g_1',g_2')$ where $g_1'(x_1)=g_1(x_1)-\nu,g_2'(x_2)=g_2(x_2).$ It follows that
	\begin{align}
		&\frac{\E_Q[(g_1(x_1)+g_2(x_2))^2]}{\E_P[(g_1(x_1)+g_2(x_2))^2]}=\frac{\E_Q[(g'_1(x_1)+g'_2(x_2))^2]+\nu^2}{\E_P[(g'_1(x_1)+g'_2(x_2))^2]+\nu^2}\\
		\le\;& \max\left\{\frac{\E_Q[(g'_1(x_1)+g'_2(x_2))^2]}{\E_P[(g'_1(x_1)+g'_2(x_2))^2]},1\right\}\le \frac{1}{1-\sigma_{\rm max}(\Sigma_{12})},
	\end{align}
	where the last inequality comes from applying case 2's result to $g'$.
	
	Combining these three cases together we prove Eq.~\eqref{equ:pf-thm-gm-1}. 
	Then Eq.~\eqref{equ:thm-gm-1} follows directly from Eq.~\eqref{equ:pf-thm-gm-1} and Lemma~\ref{lem:cov-cor-eigenvalue}.
\end{proof}

\subsection{Proof of Proposition~\ref{prop:lowerbound}}\label{app:pf-prop-lowerbound}
In this section, we prove Proposition~\ref{prop:lowerbound}.
\begin{proof}[Proof of Proposition~\ref{prop:lowerbound}]
	First, for any $t\in S^{d-1}$ and $\epsilon>0$ we construct a two-layer neural network $h_{t,\epsilon}(x)$ such that 
	\begin{enumerate}
		\item for all $x\in S^{d-1}$, $h_{t,\epsilon}(x)=0$ if $\|x-t\|_2\ge \epsilon$, 
		\item for all $x\in S^{d-1}$, $h_{t,\epsilon}(x)\ge 1$ if $\|x-t\|_2\le \epsilon/2,$ and
		\item for all $x\in S^{d-1}$, $h_{t,\epsilon}(x)\ge 0$.
	\end{enumerate}
	Recall that a two layer neural network is parameterized as $\sum_{i}a_i\relu(w_i^\top x+b_i).$
	Then $h_{t,\epsilon}(x)$ can be constructed using one neuron by setting $w_1=t, b_1=-1+\epsilon^2/2$ and $w_1=8/(3\epsilon^2).$ 
	
	We can verify the construction as follows. When $\|x-t\|_2\ge \epsilon$, we get
	\begin{align}
		\|x-t\|_2^2\ge \epsilon^2\implies 2-2x^\top t\ge \epsilon^2\implies x^\top t\le 1-\frac{\epsilon^2}{2}.
	\end{align}
	As a result, $$a_1\relu(w_1^\top x+b_1)=\frac{8}{3\epsilon^2}\relu(t^\top x-1+\epsilon^2/2)=0.$$
	
	When $\|x-t\|_2\le \epsilon/2$, we get
	\begin{align}
		\|x-t\|_2^2\le \epsilon^2/4\implies 2-2x^\top t\le \epsilon^2/4\implies x^\top t\ge 1-\frac{\epsilon^2}{8}.
	\end{align}
	As a result, $$a_1\relu(w_1^\top x+b_1)=\frac{8}{3\epsilon^2}\relu(t^\top x-1+\epsilon^2/2)\ge 1.$$
	
	Now we construct a two layer neural network $f$ such that $\|f\|_P=0$ but $\|f\|_Q=0.$ Let $X=\{x:x\in S^{d-1},\dist(x,\supp(P))\ge \epsilon\}.$ By the condition of this proposition we have $Q(X)>0.$
	
	Let $C$ be the minimum $\epsilon/2$-covering of the set $X$, and thus $|C|<\infty.$ Consider the 2-layer neural network $f(x)=\sum_{c\in C}h_{c,\epsilon}$. It follows from the definition of $X$ that $\|x-c\|_2\ge \epsilon$ for every $x\in\supp(P)$. Consequently, $f(x)=\sum_{c\in C}h_{c,\epsilon}=0$ for every $x\in\supp(P)$, which implies that $\|f\|_P=0.$ 
	
	Now for every $x\in\calX$, there exists $t\in\calC$ such that $\|x-t\|_2\le \epsilon/2.$ As a result, $h_{t,\epsilon}(x)\ge 1$, which implies that 
	\begin{align}
		f(x)=\sum_{c\in C}h_{c,\epsilon}\ge h_{t,\epsilon}(x)\ge 1.
	\end{align}
	Hence,
	\begin{align}
		\E_Q[f(x)^2]\ge \E_Q[\ind{x\in X}]=Q(X)>0.
	\end{align}

	Since $c f(x)\in\calF$ for every $c>0$, we prove the desired result.
\end{proof}
\section{Hermite Polynomial and Gaussian Kernel}
The Hermite polynomial $H_n(x)$ is an degree $n$ polynomial defined as follows,
\begin{align}
	H_n(x)=(-1)^n\exp(x^2)\frac{\dd^n \exp(-t^2)}{\dd t^n},\quad \forall n=1,2\cdots.
\end{align}
with the following orthogonality property \citep{poularikas2018handbook}:
\begin{align}
	\int \exp(-x^2/2)H_n(x/\sqrt{2})H_m(x/\sqrt{2})\dd x=\ind{n=m}2^nn!\sqrt{2\pi}.
\end{align}
As a result, we can construct a set of orthonormal basis of square integrable functions $L^2(\R)$ using the Hermite polynomial.
\begin{lemma}[see e.g., \citet{celeghini2021hermite}]\label{lem:orthonormal-basis}
	The set of functions $\{\psi_n(x)\}_{n\ge 0}$ form an orthonormal basis for $L^2(\R)$, where $$\psi_n(x)\defeq H_n\(\frac{1}{\sqrt{2}}x\)\exp\(-\frac{1}{4}x^2\)(2\pi)^{-1/4}(2^n n!)^{-1/2}.$$
\end{lemma}

The following theorem characterizes the eigenspace of a one-dimensional Gaussian kernel.
\begin{theorem}[Chapter 6.2 of \citet{fasshauer2011positive}, also see \citet{zhu1997gaussian}]\label{thm:psd-gaussian}
	For any $\epsilon>0,\alpha>0$, the eigenfunction expansion for the Gaussian $x,z\in\R$ is
	\begin{align}
		\exp\(-\epsilon^2(x-z)^2\)=\sum_{n=0}^{\infty}\lambda_n \phi_n(x)\phi_n(z),
	\end{align}
	where 
	\begin{align*}
		\lambda_n&=\frac{\alpha\epsilon^{2n}}{\(\frac{\alpha^2}{2}\(1+\sqrt{1+(2\epsilon/\alpha)^2}\)+\epsilon^2\)^{n+1/2}},\quad n=0,1,2,\cdots,\\
		\phi_n(x)&=\frac{\(1+(2\epsilon/\alpha)^2\)^{1/8}}{\sqrt{2^n n!}}\exp\(-\(\sqrt{1+(2\epsilon/\alpha)^2}-1\)\frac{\alpha^2x^2}{2}\)H_n\(\(1+(2\epsilon/\alpha)^2\)^{1/4}\alpha x\).
	\end{align*}
	And the eigenfunctions $\phi_n$ forms an orthonormal basis under weighted $L^2(\R)$ space:
	\begin{align}
		\int\phi_m(x)\phi_n(x)\frac{\alpha}{\sqrt{\pi}}\exp(-\alpha^2x^2)\dd x=\ind{m=n}.
	\end{align}
\end{theorem}

In the following theorems, we apply Theorem~\ref{thm:psd-gaussian} to the Gaussian kernels studied in this paper.
\begin{theorem}\label{thm:gaussian-kernel}
	For any $\rho\in[-1,1]$, let $P(x_1,x_2)$ be the density of $(x_1,x_2)\sim \calN\(0,\begin{pmatrix}
		1 & \rho\\\rho & 1
	\end{pmatrix}\)$ and $P(x_1),P(x_2)$ the density of corresponding marginals.
	Then the Gaussian kernel \begin{align}
		K_\rho(x_1,x_2)=&\;\frac{P(x_1,x_2)}{\sqrt{P(x_1)P(x_2)}}\\
		=\;&\frac{1}{\sqrt{2\pi(1-\rho^2)}}\exp\(-\frac{1}{2(1-\rho^2)}(x_1^2+x_2^2-2\rho x_1x_2)+\frac{x_1^2}{4}+\frac{x_2^2}{4}\)
	\end{align}
	has the following eigen-decomposition
	\begin{align}
		K_\rho(x_1,x_2)=\sum_{k=0}^{\infty}\rho^k \psi_k(x_1)\psi_k(x_2),
	\end{align}
	where $\psi_k(\cdot)$ is defined in Lemma~\ref{lem:orthonormal-basis}.
\end{theorem}
\begin{proof} We prove this theorem by considering the following two cases separately.
	
	\paragraph{Case 1: $\rho\ge 0$.} By algebraic manipulation we get
	\begin{align*}
		\sqrt{2\pi}K_\rho(x_1,x_2)=\frac{1}{(1-\rho^2)^{1/2}}\exp\(-\frac{1}{2(1-\rho^2)}\(\rho(x_1-x_2)^2+\frac{(\rho-1)^2}{2}x_1^2+\frac{(\rho-1)^2}{2}x_2^2\)\).
	\end{align*}
	Let $\epsilon^2=\frac{\rho}{2(1-\rho^2)}$ and $\alpha^2=\frac{(\rho-1)^2}{2(1-\rho^2)}$, we can equivalent write
	\begin{align}\label{equ:pf-gk-2}
		\sqrt{2\pi}K_\rho(x_1,x_2)=\frac{1}{(1-\rho^2)^{1/2}}\exp\(-\epsilon^2 (x_1-x_2)^2\)\exp\(-\frac{\alpha^2}{2}x_1^2\)\exp\(-\frac{\alpha^2}{2}x_2^2\).
	\end{align}
	By Theorem~\ref{thm:psd-gaussian} we get
	\begin{align}\label{equ:pf-gk-3}
		\exp\(-\epsilon^2 (x_1-x_2)^2\)=\sum_{n}\lambda_n\phi_n(x_1)\phi_n(x_2)
	\end{align}
	with $\int\phi_m(x)\phi_n(x)\frac{\alpha}{\sqrt{\pi}}\exp(-\alpha^2x^2)\dd x=\ind{m=n}.$
	Then we can define the function 
	\begin{align}\label{equ:pf-gk-4}
	\psi_n(x)=\(\frac{\alpha}{\sqrt{\pi}}\)^{1/2}\phi_n(x)\exp\(-\frac{\alpha^2}{2}x^2\)
	\end{align} such that $\int \psi_n(x)\psi_m(x)\dd x=\ind{m=n}.$ Combining Eqs.~\eqref{equ:pf-gk-2},\eqref{equ:pf-gk-3}, and \eqref{equ:pf-gk-4} we get
	\begin{align}
		K_\rho(x_1,x_2)=\sum_{n}\(\lambda_n\frac{1}{\sqrt{2}\alpha(1-\rho^2)^{1/2}}\)\psi_n(x_1)\psi_n(x_2).
	\end{align}
	Now we only need to prove that $\psi_n(\cdot)$ defined in Eq.~\eqref{equ:pf-gk-4} has the same form as those in Lemma~\ref{lem:orthonormal-basis}, and $\lambda_n\frac{1}{\sqrt{2}\alpha(1-\rho^2)^{1/2}}=\rho^n.$
	
	Recall that 
	\begin{align*}
		\phi_n(x)&=\frac{\(1+(2\epsilon/\alpha)^2\)^{1/8}}{\sqrt{2^n n!}}\exp\(-\(\sqrt{1+(2\epsilon/\alpha)^2}-1\)\frac{\alpha^2x^2}{2}\)H_n\(\(1+(2\epsilon/\alpha)^2\)^{1/4}\alpha x\).
	\end{align*}
	Plugin $\epsilon^2=\frac{\rho}{2(1-\rho)^2}$ and $\alpha^2=\frac{(\rho-1)^2}{2(1-\rho)^2}$ we get
	\begin{align}
		&\(1+(2\epsilon/\alpha)^2\)^{1/8}\(\frac{\alpha}{\sqrt{\pi}}\)^{1/2}=\frac{1}{\pi^{1/4}}\(\alpha^4+4\epsilon^2\alpha^2\)^{1/8}\\
		&\quad =\frac{1}{\pi^{1/4}}\(\frac{(\rho-1)^4+4\rho(\rho-1)^2}{4(1-\rho^2)^2}\)^{1/8}=\frac{1}{(2\pi)^{1/4}}\\
		&\exp\(-\frac{\alpha^2}{2}x^2\)\exp\(-\(\sqrt{1+(2\epsilon/\alpha)^2}-1\)\frac{\alpha^2x^2}{2}\)\\
		&\quad =\exp\(-\sqrt{\alpha^4+2\epsilon^2\alpha^2}\frac{x^2}{2}\)=\exp\(-\sqrt{\frac{(\rho-1)^4+4\rho(\rho-1)^2}{4(1-\rho^2)^2}}\frac{x^2}{2}\)\\
		&\quad=\exp\(-\frac{x^2}{4}\),\\
		&\(1+(2\epsilon/\alpha)^2\)^{1/4}\alpha=\(1+\frac{4\rho}{(\rho-1)^2}\)^{1/4}\(\frac{(\rho-1)^2}{2(1-\rho^2)}\)^{1/2}\\&\quad=\(\frac{(1+\rho)^2}{(\rho-1)^2}\frac{(\rho-1)^4}{(1-\rho^2)^2}\)^{1/4}\frac{1}{\sqrt{2}}=\frac{1}{\sqrt{2}}.
	\end{align}
	As a result,
	\begin{align}
		\psi_n(x)&=\(\frac{\alpha}{\sqrt{\pi}}\)^{1/2}\phi_n(x)\exp\(-\frac{\alpha^2}{2}x^2\)\\
		&=H_n\(\frac{1}{\sqrt{2}}x\)\exp\(-\frac{1}{4}x^2\)(2\pi)^{-1/4}(2^n n!)^{-1/2}.
	\end{align} 
	Now we turn to the eigenvalues. Recall that 
	$$\lambda_n=\frac{\alpha\epsilon^{2n}}{\(\frac{\alpha^2}{2}\(1+\sqrt{1+(2\epsilon/\alpha)^2}\)+\epsilon^2\)^{n+1/2}}.$$ Plugin $\epsilon^2=\frac{\rho}{2(1-\rho)^2}$ and $\alpha^2=\frac{(\rho-1)^2}{2(1-\rho)^2}$ we get
	\begin{align}
		&\frac{\alpha^2}{2}\(1+\sqrt{1+(2\epsilon/\alpha)^2}\)+\epsilon^2=\frac{1}{2}\(\alpha^2+\sqrt{\alpha^4+4\epsilon^2\alpha^2}\)+\epsilon^2\\
		&\quad =\frac{1}{2}\(\frac{(\rho-1)^2}{2(1-\rho^2)}+\frac{1}{2}\)+\frac{\rho}{2(1-\rho^2)}=\frac{1}{2(1-\rho^2)}.
	\end{align}
	Consequently, 
	\begin{align}
		\lambda_n\frac{1}{\sqrt{2}\alpha(1-\rho^2)^{1/2}}=\frac{\(\frac{\rho}{2(1-\rho^2)}\)^n}{\(\frac{1}{2(1-\rho^2)}\)^{n+1/2}}\frac{1}{\sqrt{2}(1-\rho^2)^{1/2}}=\rho^n.
	\end{align}

	\paragraph{Case 2: $\rho< 0$.} Recall that $$K_\rho(x_1,x_2)=\frac{1}{\sqrt{2\pi(1-\rho^2)}}\exp\(-\frac{1}{2(1-\rho^2)}(x_1^2+x_2^2-2\rho x_1x_2)+\frac{x_1^2}{4}+\frac{x_2^2}{4}\).$$ In this case we reuse the results from Case 1 to get
	\begin{align}
		K_{-\rho}(x_1,x_2)=\sum_{k=0}^{\infty}(-\rho)^k \psi_k(x_1)\psi_k(x_2).
	\end{align}
	By definition, we get $K_\rho(x_1,x_2)=K_{-\rho}(x_1,-x_2).$ As a result,
	\begin{align}
		K_{\rho}(x_1,x_2)=&\sum_{k=0}^{\infty}(-\rho)^k \psi_k(x_1)\psi_k(-x_2)\\
		=&\;\sum_{k=0}^{\infty}\rho^k \psi_k(x_1)\psi_k(x_2),
	\end{align}
	where the last equation follows from the fact that $\psi_k(x_2)=(-1)^k\psi_k(-x_2).$
\end{proof}

\begin{theorem}\label{thm:gaussian-kernel-multi}
	For any $\Sigma\in\R^{d_1\times d_2}$, let $P(x_1,x_2)$ be the density of $(x_1,x_2)\sim \calN\(0,\begin{pmatrix}
		I_{d_1} & \Sigma\\ \Sigma^\top & I_{d_2}
	\end{pmatrix}\)$ and $P(x_1),P(x_2)$ the density of corresponding marginals, where $x_1\in\R^{d_1},x_2\in\R^{d_2}.$
	Without loss of generality, assume $d_1\ge d_2.$ Let $\Sigma=U\Lambda V^\top$ be the singular value decomposition of $\Sigma$, where $U\in\R^{d_1\times d_2},\Lambda\in\R^{d_1\times d_2},V\in\R^{d_2\times d_2}$. Let $\sigma_1,\cdots,\sigma_{d_2}$ be the singular values of $\Sigma$ (i.e., $[\Lambda]_{i,i}=\sigma_i$). Then the Gaussian kernel \begin{align}
		K_\Sigma(x_1,x_2)=&\;\frac{P(x_1,x_2)}{\sqrt{P(x_1)P(x_2)}}
	\end{align}
	has the following eigen-decomposition
	\begin{align}
		K_\Sigma(x_1,x_2)=\prod_{i=1}^{d_2}\(\sum_{n=0}^{\infty}\sigma_i^n \psi_n([t_1]_i)\psi_n([t_2]_i)\)\prod_{i=d_2+1}^{d_1}\psi_0([t_1]_i).
	\end{align}
	where $t_1=U^\top x_1, t_2=V^\top x_2$, and $\psi_k(\cdot)$ is defined in Lemma~\ref{lem:orthonormal-basis}.
\end{theorem}
\begin{proof}
	We prove this theorem by decomposing the multi-dimensional kernel into products of several one-dimensional kernels described in Theorem~\ref{thm:gaussian-kernel}.
	
	Recall that $t_1=U^\top x_1\in\R^{d_1}, t_2=V^\top x_2\in\R^{d_2}$, where $\Sigma=U\Lambda V^\top$ is the singular value decomposition of $\Sigma.$ Let $t\in\R^{d_1+d_2}$ be the concatenation of $t_1,t_2.$ Then we have
	\begin{align}
		\E_P[tt^\top]=\begin{pmatrix}
			I_{d_1} & \Lambda\\
			\Lambda^\top & I_{d_2}
		\end{pmatrix}.
	\end{align}
	Let $P'(t_1,t_2)$ be the density of $(t_1,t_2)\sim \calN\(0,\begin{pmatrix}
		I_{d_1} & \Lambda\\ \Lambda^\top & I_{d_2}
	\end{pmatrix}\).$
	By the fact that $\Lambda$ is a diagonal matrix, we know that variables $\{[t_1]_i,[t_2]_i\}$ is independent from $\{[t_1]_j\}_{j\neq i}\cup \{[t_2]_j\}_{j\neq i}$ for every $1\le i\le d_2,$ and variable $[t_1]_i$ is independent from $\{[t_1]_j\}_{j\neq i}\cup \{[t_2]_j\}_{1\le j\le d_2}$ for every $d_2+1\le i\le d_1.$ In addition, $([t_1]_i, [t_2]_i)\sim \calN\(0,\begin{pmatrix}
		1 & \sigma_i\\ \sigma_i & 1
	\end{pmatrix}\)$. As a result, 
	\begin{align}
		K_\Sigma(x_1,x_2)=&\;\frac{P(x_1,x_2)}{\sqrt{P(x_1)P(x_2)}}=\frac{P'(t_1,t_2)}{\sqrt{P'(t_1)P'(t_2)}}\\
		=&\;\prod_{i=1}^{d_2}\(\frac{P'([t_1]_i,[t_2]_i)}{\sqrt{P'([t_1]_i)P'([t_2]_i)}}\)\prod_{i=d_2+1}^{d_1}\sqrt{P'([t_1]_i)}\\
		=&\;\prod_{i=1}^{d_2}\(\sum_{n=0}^{\infty}\sigma_i^n \psi_n([t_1]_i)\psi_n([t_2]_i)\)\prod_{i=d_2+1}^{d_1}\sqrt{P'([t_1]_i)}\label{equ:pf-gkm-1}\\
		=&\;\prod_{i=1}^{d_2}\(\sum_{n=0}^{\infty}\sigma_i^n \psi_n([t_1]_i)\psi_n([t_2]_i)\)\prod_{i=d_2+1}^{d_1}\psi_0([t_1]_i),\label{equ:pf-gkm-2}
	\end{align}
	where Eq.~\eqref{equ:pf-gkm-1} follows from Theorem~\ref{thm:gaussian-kernel-multi}, and Eq.~\eqref{equ:pf-gkm-2} follows from the definition of $\psi_0.$
\end{proof}

\section{Helper Lemmas}
\begin{lemma}\label{lem:elementwise-product-lmin}
	Let $\Sigma\in\R^{d\times d}$ be a positive semi-definite matrix with $\diag(\Sigma)=I.$ Then we have
	\begin{align}
		\forall k\ge 1,\quad \lmin(\Sigma^{\odot k})\ge \lmin(\Sigma),
	\end{align}
	where $\Sigma^{\odot k}$ denotes the element-wise $k$-th power of the matrix $\Sigma$.
\end{lemma}
\begin{proof}
	Let $\lambda=\lambda_{\min(\Sigma)}$ and we have $\Sigma-\lambda I\succeq 0.$ As a result, $(\Sigma-\lambda I)^{\odot k}=\Sigma^{\odot k}+(1-\lambda)^kI-I\succeq 0.$ Note that $0\le \lambda\le 1$ because $\Tr(\Sigma-I)=0$ implies $\lmin(\Sigma-I)\le 0$. It follows that for any $k\ge 1,$
	\begin{align}
		\Sigma^{\odot k}\succeq I-(1-\lambda)^kI\succeq \lambda I,
	\end{align}
	where the last inequality follows from the fact that $1-\lambda\ge (1-\lambda)^k$ when $k\ge 1$. Consequently,
	\begin{align}
		\lmin(\Sigma^{\odot k})\ge \lmin(\Sigma).
	\end{align}
\end{proof}

\begin{lemma}\label{lem:upperbound-singularvalue}
	Let $\Sigma\in\R^{d_1\times d_2}$ be a matrix such that 
	\begin{align}
		\begin{pmatrix}
			I_{d_1} & \Sigma\\
			\Sigma^\top & I_{d_2}
		\end{pmatrix}\succeq 0.
	\label{equ:pf-lem-us-1}
	\end{align}Then we have
	$
		\lmax(\Sigma^\top \Sigma)\le 1.
	$
\end{lemma}
\begin{proof}
	We prove by contradiction. Suppose otherwise $\lmax(\Sigma^\top \Sigma)> 1$. Let $v\in\R^{d_2}$ be the eigenvector of $\Sigma^\top \Sigma$ corresponds to its maximum eigenvalue. Then 
	\begin{align}
		\begin{pmatrix}
			-\Sigma v\\ v
		\end{pmatrix}^\top
		\begin{pmatrix}
			I_{d_1} & \Sigma\\
			\Sigma^\top & I_{d_2}
		\end{pmatrix}
		\begin{pmatrix}
			-\Sigma v\\ v
		\end{pmatrix}=
		\begin{pmatrix}
			-\Sigma v\\ v
		\end{pmatrix}^\top
		\begin{pmatrix}
			0\\ -\Sigma^\top \Sigma v+v
		\end{pmatrix}^\top=-v^\top \Sigma^\top \Sigma v + v^\top v.
	\end{align}
	By the assumption that $\lmax(\Sigma^\top \Sigma)> 1$, we get
	\begin{align}
		-v^\top \Sigma^\top \Sigma v + v^\top v=\|v\|_2^2(1-\lmax(\Sigma^\top \Sigma))<0,
	\end{align}
	which contradicts to Eq.~\eqref{equ:pf-lem-us-1}. Therefore, we must have  $\lmax(\Sigma^\top \Sigma)\le 1.$
\end{proof}

\begin{lemma}\label{lem:cov-cor-eigenvalue}
	Let $M$ be a positive semi-definite matrix with the following form
	\begin{align}
		M=\begin{pmatrix}
			I_{d_1} & \Sigma\\
			\Sigma^\top & I_{d_2}
		\end{pmatrix}
	\end{align}
	for some $d_1,d_2>0$. Then we have
	\begin{align}
		\lmin(M)=1-\sigma_{\rm max}(\Sigma),
	\end{align}
	where $\sigma_{\rm max}$ is the largest singular value of $\Sigma$.
\end{lemma}
\begin{proof}
	Without loss of generality, we assume $d_1\ge d_2.$ Let $\Sigma=U\Lambda V^\top$ be the singular value decomposition of $\Sigma$, with $U\in\R^{d_1\times d_1},V\in\R^{d_2\times d_2},\Lambda\in\R^{d_1\times d_2}.$ Then we have
	\begin{align}
		M=
		\begin{pmatrix}
			U & 0\\
			0 & V
		\end{pmatrix}
		\begin{pmatrix}
			I_{d_1} & \Lambda\\
			\Lambda^\top & I_{d_2}
		\end{pmatrix}
		\begin{pmatrix}
			U^\top & 0\\
			0 & V^\top
		\end{pmatrix}.
	\end{align}
	Let $\bar{M}=\begin{pmatrix}
		I_{d_1} & \Lambda\\
		\Lambda^\top & I_{d_2}
	\end{pmatrix}$. Because $U,V$ are orthonormal matrices, we get $\lmin(M)=\lmin(\bar{M}).$ Note that $[\Lambda]_{i,j}=0$ whenever $i\neq j$ and $[\Lambda]_{i,i}=\sigma_i(\Sigma)$. Consequently, the eigenvalues of $\bar{M}$ is
	\begin{align}
		1\pm \sigma_1(\Sigma),1\pm \sigma_2(\Sigma),\cdots,1\pm \sigma_{d_2}(\Sigma)
	\end{align} with multiplicity 1, and $1$ with multiplicity $d_1-d_2.$ As a result,
	$$\lmin(M)=\lmin(\bar{M})1-\sigma_{\rm max}(\Sigma).$$
\end{proof}
\section{Implementation Details}\label{app:experiments}
\paragraph{The synthetic dataset.} The features to the dataset is $x=(x_1,x_2)$ where $x_1\sim \mathbb{R}^{d_1},x_2\sim \mathbb{R}^{d_2}$. We set $d_1=d_2=256.$ The ground-truth labeling function is $f^\star(x)=f_1^\star(x_1)+f_2^\star(x_2)$, where $f_1^\star,f_2^\star$ are two layer neural networks with ReLU activation, hidden dimension 128, and random initialization. For simplicity, labels are noiseless and we draw fresh samples on every iteration during the training process.

For both the ID and OOD distribution, the feature vector $x$ has Gaussian distribution with covariance $\Sigma_P,\Sigma_Q$. To match the setting of Theorem~\ref{thm:gaussian-multi}, the covariance $\Sigma_P,\Sigma_Q$ are generated in the following way:
\begin{align}
	\Sigma_P=\begin{pmatrix}
		I & \gamma O_P\\ \gamma O_P^\top & I
	\end{pmatrix},\Sigma_Q=\begin{pmatrix}
	I & \gamma O_Q\\ \gamma O_Q^\top & I
\end{pmatrix},
\end{align}
where $O_P,O_Q$ are two random orthonormal matrices, and $\gamma=0.9.$ Consequently, $\Sigma_P\succeq \Omega(1)\Sigma_Q.$

\paragraph{The model classes.}
The structured model is of the form $f_1(x_1)+f_2(x_2)$, where $f_1,f_2$ are two-layer neural networks with hidden dimension 512 and ReLU activation. The parameters of $f_1,f_2$ are initialized by $\mathcal{N}(0, \sigma^{2})$ with $\sigma=10^{-3}.$ We train the model with batch size 1024, 500 batches per epoch, and 200 epochs in total. We use SGD with learning rate $3*10^{-3}$ and momentum $0.9$.

The unstructured model is a two layer neural network on the concatenation of $x_1$ and $x_2$ with hidden dimension 1024 and ReLU activation. Therefore, the unstructured model is a superset of the structured model (with hidden dimension 512). We train the unstructured model for 235 epochs to get a similar ID loss.\footnote{Because the unstructured model has more parameters, it trains slower than the structured model.} We keep other hyper-parameters  exactly the same as the structured model case.

\subsection{Ablation Study.}

\paragraph{The choice of hidden dimensions.}
We first vary the hidden dimensions of the unstructured model to test whether the differences of structured and unstructured model come from the number of parameters. Table~\ref{tab:2} summarizes the result---with a smaller model, the ground-truth labeling function cannot be expressed by the model class (since the ID loss doesn't converge to 0); with a larger model, the OOD loss becomes larger.

\begin{table}[htp]
	\begin{center}
		\begin{tabular}{ |c|c|c| } 
			\hline
			Model Class & ID loss & OOD loss \\ 
			\hline 
			Unstructured model $f(x)$, hidden dim = 256 & 0.26957 & 0.8759 \\ 
			Unstructured model $f(x)$, hidden dim = 512 & 0.01688 &	0.07774 \\ 
			Unstructured model $f(x)$, hidden dim = 1024 & 0.00105	& 0.01235 \\ 
			Unstructured model $f(x)$, hidden dim = 2048 & 0.00129 & 0.01395 \\ 
			\hline
		\end{tabular}
	\end{center}
	\caption{The ID and OOD losses of unstructured models with different size.}
	\label{tab:2}
\end{table}

\paragraph{The effect of regularization.}
We also add $\ell_1$ and $\ell_2$ regularization to the unstructured model, since the structured model is $1/2$ sparse when viewed as a unstructured model. However, even though the regularization helps OOD loss to some extent, the OOD loss of the unstructured model is still far worse than the OOD loss of the structured model (Table~\ref{tab:3}).

\begin{table}[htp]
	\begin{center}
		\begin{tabular}{ |c|c|c| } 
			\hline
			Model Class & ID loss & OOD loss \\ 
			\hline 
			Structured model $f_1(x_1)+f_2(x_2)$, no regularization & 0.00100 & 0.00118 \\ 
			Unstructured model $f(x)$, $\ell_1$ regularization & 0.00116 & 0.00899\\
			Unstructured model $f(x)$, $\ell_2$ regularization & 0.00142 &	 0.01623\\
			\hline
		\end{tabular}
	\end{center}
	\caption{The ID and OOD losses of unstructured models with different regularizations.}
	\label{tab:3}
\end{table}

\end{document}